\documentclass[runningheads]{llncs}
\usepackage{graphicx}
\usepackage[table]{xcolor}

\usepackage{xspace}
\usepackage{wrapfig}
\makeatletter
\DeclareRobustCommand\onedot{\futurelet\@let@token\@onedot}
\def\@onedot{\ifx\@let@token.\else.\null\fi\xspace}

\def\eg{\emph{e.g}\onedot} 
\def\ie{\emph{i.e}\onedot} 
\def\cf{\emph{c.f}\onedot}

\makeatother
\usepackage{tikz}
\usepackage{comment}
\usepackage{amsmath,amssymb} %

\usepackage[accsupp]{axessibility}  %

\usepackage[normalem]{ulem}
\usepackage{tikz}
\usetikzlibrary{patterns}

\usepackage{tabularx}
\usepackage{epsfig}
\usepackage{stmaryrd}
\usepackage{graphicx}
\usepackage[utf8]{inputenc}
\usepackage{amsmath}
\usepackage{amssymb}
\usepackage{enumitem}
\usepackage{booktabs}
\usepackage{textcomp}
\usepackage{gensymb}
\usepackage{multirow}
\usepackage{caption}
\usepackage{breakcites}
\usepackage{subcaption}
\usepackage{makecell}
\usepackage{comment}
\usepackage{soul}
\usepackage{xspace}
\usepackage[export]{adjustbox}
\usepackage{microtype}
\usepackage{colortbl}
\usepackage[pagebackref,breaklinks,colorlinks]{hyperref}

\usepackage[capitalize]{cleveref}
\crefname{section}{Sec.}{Secs.}
\Crefname{section}{Section}{Sections}
\Crefname{table}{Table}{Tables}
\crefname{table}{Tab.}{Tabs.}

\newcommand{\name}{ManiFest\xspace}
\newcolumntype{H}{>{\setbox0=\hbox\bgroup}c<{\egroup}@{}}

\DeclareRobustCommand{\ulred}[1]{\setulcolor{red}\ul{#1}}
\soulregister{\ulred}{1}
\DeclareRobustCommand{\ulgreen}[1]{\setulcolor{green}\ul{#1}}
\soulregister{\ulgreen}{1}

\definecolor{myGermG}{rgb}{0.270,0.466,0.749}
\definecolor{myGermE}{rgb}{0.839,0.677,0.190}
\newcommand{\germGcol}[1]{\textcolor{myGermG}{#1}}
\newcommand{\germEcol}[1]{\textcolor{myGermE}{#1}}

\newcommand\tikzmark[2]{%
\tikz[remember picture,baseline] \node[above, outer sep=0pt, inner sep=0pt] (#1){\phantom{#2}};%
}

\newcommand\linkwithlabel[3]{%
	\begin{tikzpicture}[remember picture, overlay, >=stealth, shift={(0,0)}]
		\draw[thick,->] (#1) -- (#2) node [midway,fill=white,color=white,text=black] {#3};
	\end{tikzpicture}%
}

\begin{document}
\pagestyle{headings}
\mainmatter
\def\ECCVSubNumber{6959}  %

\title{\name: manifold deformation \\ for few-shot image translation} %

\titlerunning{ManiFest: manifold deformation for few-shot image translation} 
\authorrunning{Fabio Pizzati et al.} 
\author{Fabio Pizzati$^{1,2}$, Jean-François Lalonde$^3$, Raoul de Charette$^1$}
\institute{$^1$Inria, $^2$VisLab, $^3$Université Laval\\
\email{\{fabio.pizzati, raoul.de-charette\}@inria.fr, jflalonde@gel.ulaval.ca}}
\index{de Charette, Raoul}

\maketitle

\begin{abstract}
Most image-to-image translation methods require a large number of training images, which restricts their applicability.  
We instead propose ManiFest: a framework for few-shot image translation that learns a context-aware representation of a target domain from a few images only. 
To enforce feature consistency, our framework learns a style manifold between source and additional anchor domains (assumed to be composed of large numbers of images). The learned manifold is interpolated and deformed towards the few-shot target domain via patch-based adversarial and feature statistics alignment losses. All of these components are trained simultaneously during a single end-to-end loop. 
In addition to the general few-shot translation task, our approach can alternatively be conditioned on a single exemplar image to reproduce its specific style. Extensive experiments demonstrate the efficacy of ManiFest on multiple tasks, outperforming the state-of-the-art on all metrics. Our code is avaliable at \url{https://github.com/cv-rits/ManiFest}.
\end{abstract}

\begin{figure}[t]
    \centering
    \includegraphics[width=0.8\linewidth]{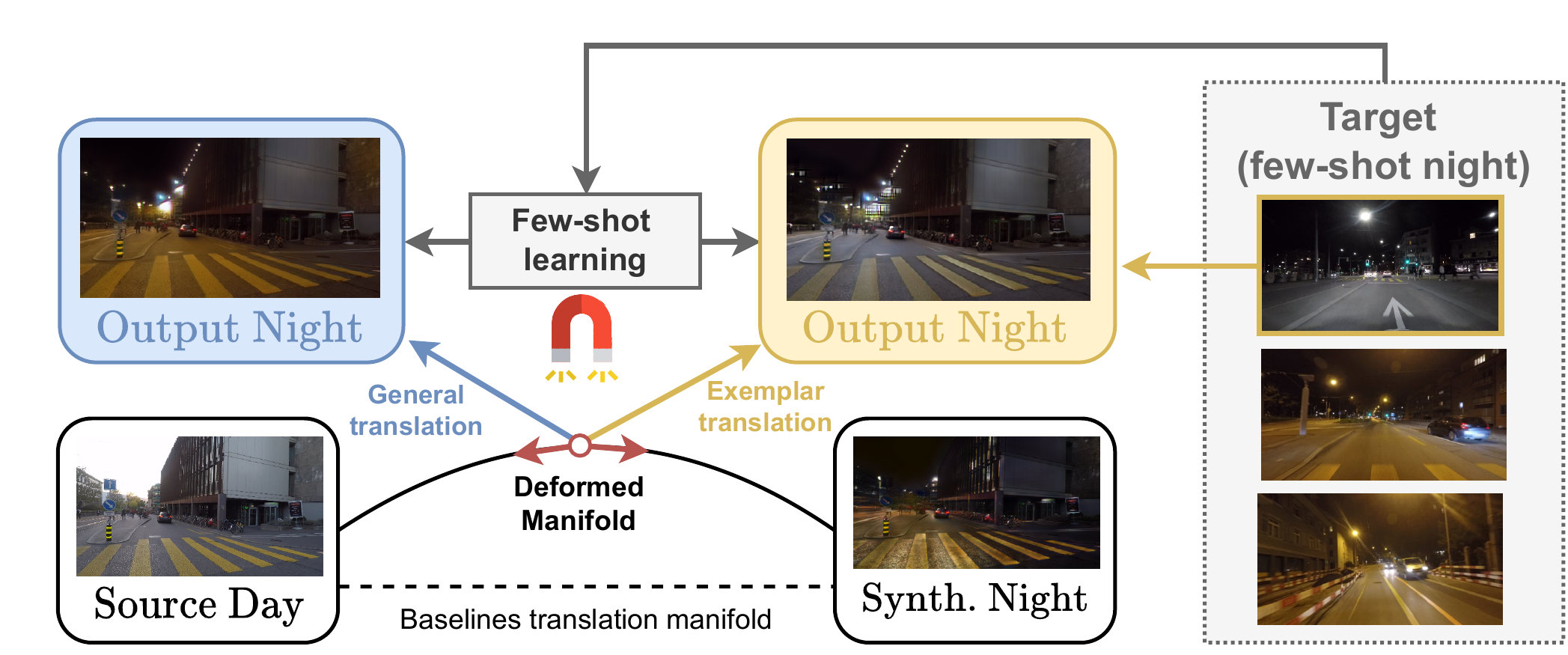}
    \caption{Overview of \name, which translates images from a source domain (here, Day) to a few-shot target (Night). Our framework learns a manifold between \emph{anchor} domains, in the example spanning the translation between Day and Synthetic Nighttime. Our system deforms the manifold by injecting the few-shot domain information between anchor style representations, and further departs from the deformed manifold by learning to approximate the target domain \germGcol{\textit{general}} appearance, or to reproduce the style of a particular \germEcol{\textit{exemplar}}.}
    \label{fig:teaser}
\end{figure}

\section{Introduction}

Image-to-image translation (i2i) frameworks are gaining traction on multiple applications such as autonomous driving~\cite{li2019bidirectional,sakaridis2020map} as well as photo editing~\cite{park2019semantic,park2020swapping}. Those methods rely on the availability of large-scale datasets, and as such they are restricted to applications where large quantities of images are available.\\
Unfortunately, it is unrealistic to impose significant data collection constraints every time a new i2i scenario is pursued. In addition to the complex logistics involved in acquiring large quantities of images, some scenarios may be rare (\eg, auroras) or dangerous (\eg, erupting volcanoes) thereby preventing even the capture of sufficient training data. Existing methods have been proposed to alleviate the requirement for large datasets, but they mostly show realistic results in highly structured environments such as face translation~\cite{liu2019few,saito2020coco,li2021super}.\\
In this context, we propose \name, a framework for few-shot image-to-image translation which is shown to be robust to highly unstructured transformations such as adverse weather generation or night rendering. Our approach, illustrated in \cref{fig:teaser}, starts from the observation that features consistency (\ie which image parts should be translated together) is crucial for unstructured i2i~\cite{ma2018exemplar} and that the few-shot domain offers little cues to train efficiently without overfitting~\cite{ojha2021few}. Indeed, rather than directly addressing few-shot i2i, \name exploits features learned on a stable manifold for the few-shot domain transformation. To do so, it leverages techniques inspired from style transfer and patch-based training. 
We learn either to translate to some \textit{general} style approximating the entire few-shot set, or to reproduce a specific \textit{exemplar} from it.
In short, our contributions are:
\begin{itemize}[noitemsep,topsep=0pt]
    \item \name, a few-shot image translation framework using feature consistency~by weighted manifold interpolation (WMI) and local-global few-shot loss (LGFS).
    \item We introduce GERM, a novel residual correction mechanism for enabling general and exemplar translation, that also boosts performances.
    \item Our framework outperforms previous work on adverse weather and low-light few-shot image translation tasks. We also present qualitative evaluations on rare (auroras) and dangerous (volcanoes) events. 
\end{itemize}
We discuss related works in Sec.~\ref{sec:related} and present our method in Sec.~\ref{sec:method}. The latter is thoroughly evaluated in Sec.~\ref{sec:experiments} and we show several extensions in Sec.~\ref{sec:extensions}.

\section{Related work}
\label{sec:related}

\noindent \textbf{Image-to-image translation (i2i)} \quad
Although the early i2i translation methods required paired data~\cite{isola2017image,zhu2017toward}, cycle-consistency~\cite{zhu2017unpaired,liu2017unsupervised} or recent alternatives with contrastive learning~\cite{park2020contrastive} have lifted such constraint. 
Many approaches separate style and content to enable multi-modal or multi-target translations~\cite{huang2018multimodal,lee2019drit++,jiang2020tsit,choi2020stargan,yu2019multi}, while others use additional strategies to increase scene contextual preservation~\cite{zheng2021spatially,jia2021semantically}. 
Translation networks can be conditioned on a variety of additional information, including semantics~\cite{li2018semantic,ramirez2018exploiting,tang2020multi,cherian2019sem,zhu2020semantically,zhu2020sean,lin2020multimodal,ma2018exemplar}, instances~\cite{mo2018instagan}, geometry~\cite{wu2019transgaga}, models~\cite{pizzati2021guided,pizzati2021comogan,pizzati2020model,hu2021model,tremblay2020rain}, low-resolution inputs~\cite{abid2021image} or exemplar images~\cite{ma2018exemplar,zhang2020cross,zhan2021unbalanced}. Still, all require a large amount of data.

\noindent \textbf{GANs with limited data} \quad
There have been several attempts to overcome the large data requirement for training GANs. Some use transfer learning~\cite{torrey2010transfer} to adapt previously-trained networks to new few-shot tasks~\cite{wang2021transferi2i,li2020few}. In particular,~\cite{ojha2021few} uses a patch-based discriminator to generalize to few-shot domains. However, these methods are designed for generative networks and do not immediately apply to i2i. Another line of work focuses on the limited data scenario~\cite{patashnik2021balagan,cao2021remix,zhao2020differentiable,karras2020training}, but usually performs poorly when very few (10--15) images are used for training. Others exploit additional knowledge to enable few-shot or zero-shot learning, such as pose-appearance decomposition~\cite{wang2020semi}, image conditioning~\cite{endo2021fewshotsmis} or textual inputs~\cite{lin2021zstgan}. FUNIT~\cite{liu2019few} and COCO-FUNIT~\cite{saito2020coco} use few-shot style encoders to adapt the network behavior at inference time. Some use meta-learning to adapt quickly to newly seen domains~\cite{lin2019learning}. Those methods show limited performance on highly unstructured scenarios.~\cite{dell2021leveraging} exploits geometry for patch-based few-shot training, but only on limited domains with specific characteristics.

\noindent \textbf{Neural style transfer} \quad
Style transfer could be seen as an instance of few-shot i2i, where the goal is combining content and style of two images~\cite{gatys2016image}. This may result in distortions, which some work tried to mitigate~\cite{luan2017deep}. The first examples of style transfer with arbitrary input styles are in~\cite{huang2017arbitrary,li2017universal}. Others try to transfer styles in a photo-realistic manner by using a smoothing step~\cite{li2018closed} or with wavelet transforms~\cite{yoo2019photorealistic}. These methods provide good results in some controlled scenarios, but they may fail to understand contextual mappings between source and style elements (\eg sky, buildings, etc.) which we learn accurately.

\section{\name{}}
\label{sec:method}

\begin{figure*}[t]
    \centering
    \includegraphics[width=\linewidth]{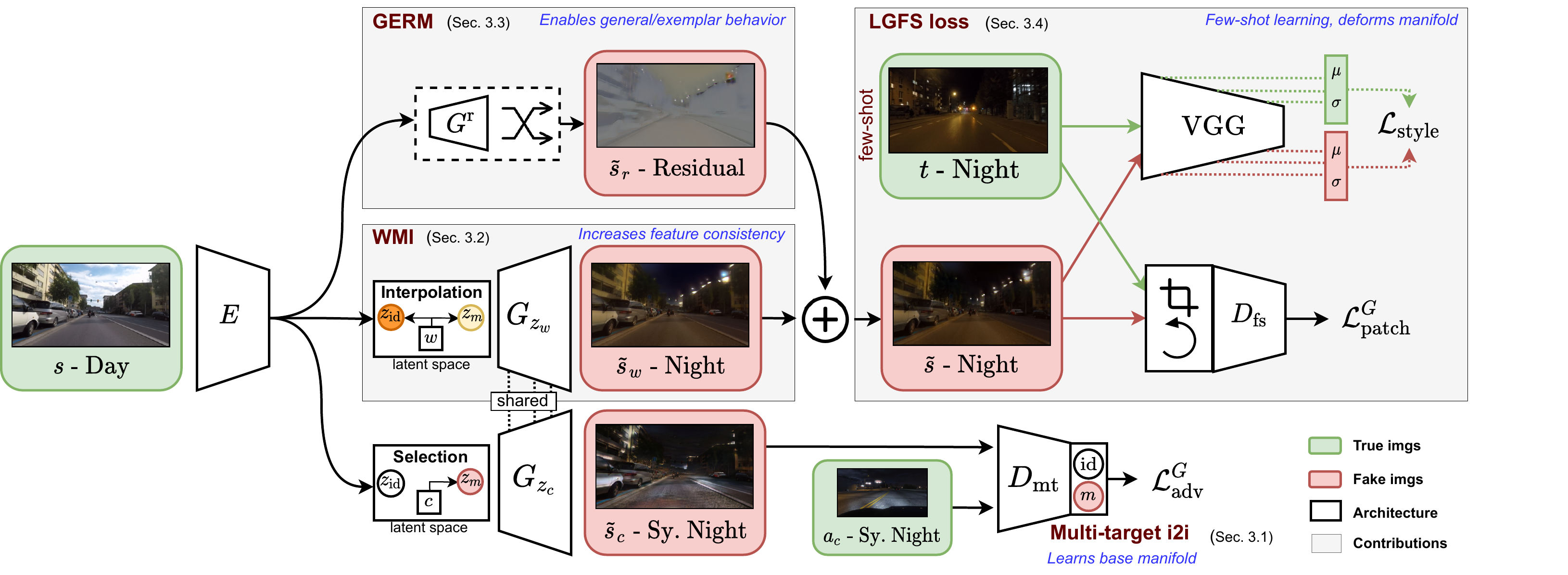}
    \caption{\name architecture, here translating $\text{Day} \mapsto \text{Night}$ using few real night images and a synthetic night anchor domain. 
    	The encoded image representation $E(s)$ is separated into content and style codes, and translated to the few-shot domain by injecting $\mathcal{T}$ on a manifold learned on anchor domains in a multi-target i2i setting (bottom). %
    	We correct the output by using residuals estimated by the GERM (top). The LGFS loss (top-right), based on statistics alignment and patch-based adversarial learning, deforms the manifold and injects $\mathcal{T}$ in it. 
        The reconstruction cycle with a style encoder is omitted for simplicity and follows~\cite{huang2018multimodal}.
        }
    \label{fig:architecture}
\end{figure*}

The few-shot i2i task consists in learning a $\mathcal{S}\mapsto\mathcal{T}$ mapping between images of a source domain $\mathcal{S}$ and a target domain $\mathcal{T}$ containing few training samples (\eg, $|\mathcal{T}| \leq 25$). Fig.~\ref{fig:architecture} presents an overview of our approach. 
We learn a style manifold in a standard multi-target GAN fashion (Sec.~\ref{sec:mt-i2i}) from a set of domains which contain large amounts of training data. We call these domains \emph{anchors}, and denote them $\mathbb{A}$. 
The idea of \name is to simultaneously 1) learn a stable manifold using anchor domains and 2) perform few-shot training by enforcing the target style appearance to lie within the learned manifold. This allows to exploit additional knowledge, like feature consistency (\ie, image parts to be translated together), learned on anchors.
To this end,  Weighted Manifold Interpolation (WMI, Sec.~\ref{sec:wmi}) exploits style interpolation to benefit from the learned feature consistency on anchors. We allow to further depart from the interpolated manifold with the General-Exemplar Residual Module (GERM, Sec.~\ref{sec:germ}) which learns a residual image refining the overall appearance and thus enabling style transfer to the \emph{general} few-shot style (approximating the entire set $\mathcal{T}$), or to a single \emph{exemplar} in $\mathcal{T}$ as in~\cite{ma2018exemplar}.
We learn the appearance of $\mathcal{T}$ and inject it in the manifold with the Local-Global Few-Shot loss (LGFS, Sec.~\ref{sec:lgfs}).
In the following, \textit{real} images are $s\in\mathcal{S},t\in\mathcal{T}$, and \textit{fake} ones $\tilde{s}\in\mathcal{T}$ where $\tilde{s}$ is our output. 

\subsection{Multi-target i2i}
\label{sec:mt-i2i}

Instead of learning $\mathcal{S}\mapsto\mathcal{T}$ directly, we assume the availability of a set of two \emph{anchor} domains, ${\mathbb{A} = \{\mathcal{A}_\mathrm{id}, \mathcal{A}_m\}}$, with abundant data (equivalent to the ``base'' categories in few-shot classification, \eg, \cite{chen2019closer}). By construction, one anchor is always the identity domain ($\mathcal{A}_\mathrm{id} = \mathcal{S}$), while the other ($\mathcal{A}_m$) contains images easier to collect than $\mathcal{T}$, for example synthetic images or images from existing datasets. We formalize the multi-target image translation problem as learning the $\mathcal{S}\mapsto \mathbb{A}$ mapping. At training time, we disentangle image content and appearance by using content and style encoders $E(\cdot)$ and $Z(\cdot)$ respectively. We use $Z$ for reconstruction and translation as in~\cite{huang2018multimodal}, which we refer for details. We reconstruct $s=G_{Z(s)}(E(s))$, where $G_{Z(s)}$ is the style injection of $Z(s)$ into $G$ as in~\cite{huang2018multimodal}. This effectively learns latent style distributions as in~\cite{huang2018multimodal}, namely here for each anchor $\{z_{\text{id}}, z_{\text{m}}\}$. A multi-target mechanism (following \cite{choi2020stargan}) is employed in $Z$ since we have two anchors. We translate to a randomly selected domain $c\in\{\text{id}, m\}$ with
\begin{equation}
        z_c = \llbracket c = \text{id} \rrbracket z_{\text{id}} + \llbracket c = m \rrbracket z_{m} \,, \quad 
        \tilde{s}_c = G_{z_c}(E(s)) \,,
\end{equation} where $\llbracket \cdot \rrbracket$ are the Iverson brackets. The translation to a given anchor style is depicted in Fig.~\ref{fig:architecture} as ``\textbf{selection}''.
The multi-target discriminator $D_{\mathrm{mt}}$ employs adversarial losses $\mathcal{L}_{\mathrm{adv}}^G$ and $\mathcal{L}_{\mathrm{adv}}^D$ to force fake images $\tilde{s}_c$ to resemble $a_c \in \mathcal{A}_c$. Additional training details are in the supp. material.

\subsection{Weighted Manifold Interpolation (WMI)}
\label{sec:wmi}

Our intuition is that encoding $\mathcal{T}$ between the linearly interpolated style representations of $\mathbb{A}$ should enforce feature consistency in $\mathcal{T}$. 
For instance, assuming $\mathcal{S}$ = \textit{day}, $\mathcal{T}$ = \textit{night}, $\mathcal{A}_m$ = \textit{synthetic night}, the network will be provided with the information that all sky pixels should be darkened together.

In practice, we learn weights $w=\{w_{\text{id}}, w_m\}$ which sum to 1 and encode an image $\tilde{s}_w$ with feature consistency by interpolating the anchors style representations:
\begin{equation}
    z_w = w_{\text{id}}z_{\text{id}} + w_m z_m \,, \quad
    \tilde{s}_w = G_{z_w}(E(s)) \,.
\end{equation}
This is visualized in Fig.~\ref{fig:architecture} as \textbf{``interpolation''}. Learning $w$ allows us to determine the point in the $\mathbb{A}$ manifold which is most consistent with $\mathcal{T}$. This point is learned with the LGFS loss (Sec.~\ref{sec:lgfs}).

\begin{figure}[t]
    \hspace{2cm}
    \includegraphics[width=0.7\linewidth]{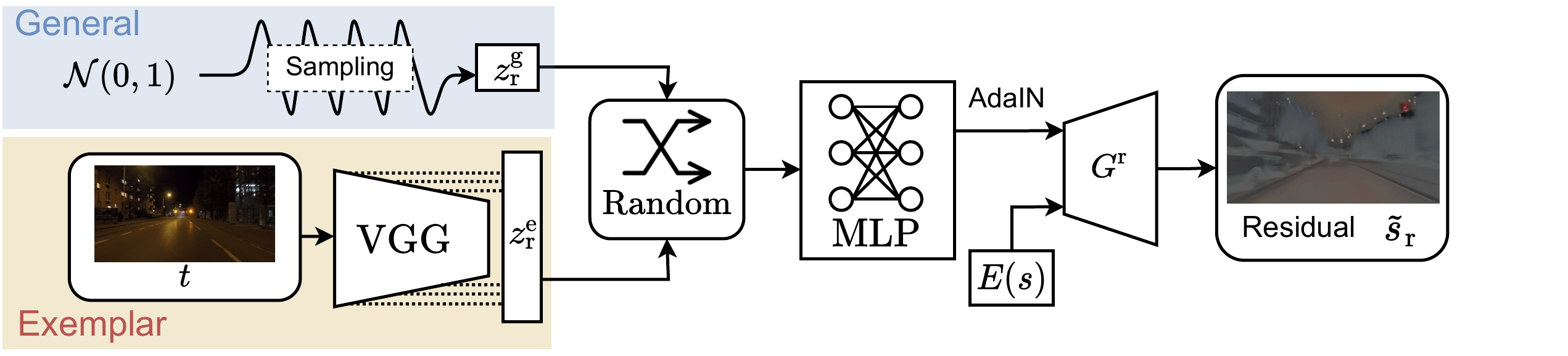}
    \caption{GERM-based residuals. We perform either \germEcol{\textit{exemplar}}- or \germGcol{\textit{general}}-based transformations on the few-shot set by learning residuals conditioned on original image features ($E(s)$) and extracted statistics or noise, respectively. At training time, we alternate randomly the two modalities.}
    \label{fig:residuals}
\end{figure}

\subsection{General-Exemplar Residual Module (GERM)}
\label{sec:germ}
Our GERM seeks to further increase realism by learning a residual in image space. Moreover, our design enables distinguishing between \textit{general} and \textit{exemplar} translations. The idea is to allow deviations from the $\mathbb{A}$ manifold by learning a residual image $\tilde{s}_\mathrm{r}$ which helps encode missing characteristics from $\mathcal{T}$.
This is done by processing the input image features $E(s)$ with a generator $G^\mathrm{r}$ such that
\begin{equation}
    \tilde{s}_\mathrm{r} = G^\mathrm{r}_{z_\mathrm{r}}(E(s)) \,, \quad \text{and} \quad
    \tilde{s} = \tilde{s}_{w} + \tilde{s}_\mathrm{r} \,,
    \label{eqn:residual}
\end{equation}
where $z_\mathrm{r}$ is a vector controlling general- or exemplar-based modalities. In both cases, we draw inspiration from AdaIN style injection~\cite{huang2018multimodal} and condition the injected parameters on different vectors, as illustrated in Fig.~\ref{fig:residuals}.
	
For the \textit{exemplar} residual, the style of a specific image $t\in\mathcal{T}$ as in~\cite{ma2018exemplar} is reproduced by conditioning the residual on $t$. In this case, 
\begin{equation}
    z^\mathrm{e}_\mathrm{r} = (\mu_k(t), \sigma_k(t))\mathbin|_{k=1}^K \,, \quad
    \tilde{s}_\mathrm{r} = G^\mathrm{r}_{z^\mathrm{e}_\mathrm{r}}(E(s)) \,,
\end{equation}
where $\mu_k(\cdot) = \mu(\phi_k(\cdot))$ and $\sigma_k(\cdot) = \sigma(\phi_k(\cdot))$ are mean and variance of the $k$-th out of $K$ layer outputs $\phi_k$ of a pretrained VGG network~\cite{huang2017arbitrary}, and $|$ is the concatenation operator. Since the LGFS loss exploits VGG statistics (Sec.~\ref{sec:lgfs}), $G^\mathrm{r}$ will be driven to exploit the additional information provided by the input statistics vector, effectively making the generated image more similar to $t$.

We learn a \textit{general} residual by removing the conditioning on $t$ and by injecting random noise instead:
\begin{equation}
    z^\mathrm{g}_\mathrm{r} \sim \mathcal{N}(0, 1) \,, \quad 
    \tilde{s}_\mathrm{r} = G^\mathrm{r}_{z^\mathrm{g}_\mathrm{r}}(E(s)) \,.
\end{equation}

\subsection{Local-Global Few-Shot loss (LGFS)}
\label{sec:lgfs}

To guide the learning, the resulting image $\tilde{s}$ is compared against the few-shot training set $\mathcal{T}$ with a combination of two loss functions. 
First, we take inspiration from the state-of-the-art of image style transfer where one image is enough for transferring the \textit{global} appearance of the style scene~\cite{huang2017arbitrary}. Our intuition is that feature statistics alignment, widely used in style transfer, could be less prone to overfitting with respect to adversarial training. Therefore, we align features between $\tilde{s}$ and a target image $t \in \mathcal{T}$ using style loss $\mathcal{L}_{\mathrm{style}}$ as in \cite{huang2017arbitrary}
\begin{equation}
    \mathcal{L}_{\mathrm{style}} = \sum^{K}_{k=1}||\mu_k(\tilde{s}) - \mu_k(t)||_2 + ||\sigma_k(\tilde{s}) - \sigma_k(t)||_2 \, ,
\end{equation}
where $(\mu_k, \sigma_k)$ are the same as in Sec.~\ref{sec:germ}. While this is effective in modifying the general image appearance, aligning statistics alone is insufficient to produce realistic outputs. Thus, to provide \textit{local} guidance, \ie, on more fine-grained characteristics, we employ an additional discriminator $D_{\mathrm{fs}}$ which is trained to distinguish between rotated patches sampled from $\tilde{s}$ and $t$. We define the adversarial losses~\cite{mao2017least}:
\begin{equation}
\begin{split}
\mathcal{L}_{\mathrm{patch}}^G &= ||D_{\mathrm{fs}}(p(\tilde{s})) - 1||_2 \,,\\
\mathcal{L}_{\mathrm{patch}}^{D} &= ||D_{\mathrm{fs}}(p(\tilde{s}))||_2 + ||D_{\mathrm{fs}}(p(t)) - 1||_2 \,,
\end{split}
\end{equation}
where $p$ is a random cropping and rotation function. Note how the \textit{exemplar} residual (from Sec.~\ref{sec:germ}) is conditioned on the \emph{same} feature statistics used here---this is what enables the exemplar-based behavior of the network. Also note the interaction between components: backpropagating the LGFS loss \textit{deforms} the manifold learned by multi-target i2i, at the point identified by WMI, thereby injecting $\mathcal{T}$ ``between'' $\{\mathcal{A}_\text{id},\mathcal{A}_m\}$. We provide additional visualizations of the deformed manifold in the supp. video and material.

\subsection{Training strategy}
\label{sec:training}

Our framework is fully trained end-to-end and optimizes
\begin{equation}
	\min_{\Theta(E,G,G^r,Z),w} \mathcal{L}_{\mathrm{style}} + \mathcal{L}^G_{\mathrm{patch}} + \mathcal{L}^G_{\mathrm{adv}} \quad \text{and} 
	\min_{\Theta(D_{\mathrm{fs}},D_{\mathrm{mt}})} \mathcal{L}^D_{\mathrm{patch}} + \mathcal{L}^D_{\mathrm{adv}} \,,\\
\end{equation}
where $\Theta(\cdot)$ refers to the network parameters. 
We train GERM (Sec.~\ref{sec:germ}) by randomly selecting one of the exemplar or general mode at each training iteration. For the multi-target settings, we adapt the discriminator and the style encoder of our backbone in a multi-target setup following~\cite{choi2020stargan}.

\section{Experiments}\label{sec:experiments}

We leverage 4 datasets~\cite{sakaridis2021acdc,sakaridis2020map,cordts2016cityscapes,richter2017playing} and 3 translation tasks (Sec.~\ref{sec:tasks}) and evaluate performances against recent baselines \cite{liu2019few,saito2020coco,yoo2019photorealistic,ma2018exemplar,huang2018multimodal} (Sec.~\ref{sec:sota-comparison}). We further demonstrate the benefit of our few-shot translation on a downstream segmentation task (Sec.~\ref{sec:exp-seg}), and rare few-shot scenarios (Sec.~\ref{sec:exp-volcanoes}), and finally ablate our contributions (Sec.~\ref{sec:ablation}). 
In all, we use MUNIT~\cite{huang2018multimodal} as our backbone.

\subsection{Training setup}

\subsubsection{Datasets}
\label{sec:datasets}
We use four datasets for our experiments.\\
\textbf{ACDC \:} We use ACDC~\cite{sakaridis2021acdc} for most of our experiments, using the night\slash rain\slash snow\slash fog conditions with 400/100/500 images for train/val/test respectively, following official splits. For any individual condition, ACDC also includes geolocalized weakly-paired clear weather day images of same splits.\\
\textbf{Dark Zurich \:} Similar to ACDC, Dark Zurich (DZ)~\cite{sakaridis2020map} has daytime images paired with nighttime/twilight conditions. Here, we focus on twilight conditions exclusively and use training images from the GOPRO348 sequence only since it exhibits a distinctive twilight appearance. We split the total 819 image pairs into 25/794 for train/test, respectively.\\
\textbf{Cityscapes \:} Cityscapes~\cite{cordts2016cityscapes} is used to evaluate \name for training segmentation networks robust to nighttime\footnote{ACDC does not provide annotated daytime clear weather sequences.}. It includes 2975/500/1525 annotated images for train/val/test.\\
\textbf{VIPER \:} As anchors, we employ synthetic images from the VIPER dataset~\cite{richter2017playing}, using the condition metadata to define splits. 4137/3090/1305/2018/2817 images are extracted from the VIPER training set for day/night/rain/snow/sunset conditions, respectively.

\subsubsection{Tasks and evaluation}
\label{sec:tasks}

We train our framework on three main tasks: \\
\textbf{$\text{Day}\mapsto\text{Night}$ \:} on ACDC daytime ($\mathcal{S}$) and nighttime ($\mathcal{T}$). \\
\textbf{$\text{Clear}\mapsto\text{Fog}$ \:} on ACDC daytime ($\mathcal{S}$) and fog ($\mathcal{T}$). \\
\textbf{$\text{Day}\mapsto\text{Twilight}$ \:} on DZ daytime ($\mathcal{S}$) and twilight ($\mathcal{T}$).\\ %
Unless mentioned otherwise, the (synthetic) anchor domains from VIPER are ``night'' for $\text{Day}\mapsto\text{Night}$ and $\text{Day}\mapsto\text{Twilight}$, and ``day'' for $\text{Clear}\mapsto\text{Fog}$. We evaluate with the FID~\cite{heusel2017gans} and LPIPS~\cite{zhang2018unreasonable} metrics. 
While FID compares feature distance globally, LPIPS compares translated source images and the geolocalized paired image in the target dataset. This is beneficial for evaluating our exemplar modality. For all, we train on downsampled x4 images.

\begin{figure}[t!]
	\centering
	\begin{subfigure}{0.53\linewidth}
		\resizebox{\linewidth}{!}{
			\setlength{\tabcolsep}{0.003\linewidth}
            \tiny
			\begin{tabular}{c c c c c}
				\toprule
				&\adjustbox{valign=m}{\rotatebox{90}{Anchor}}
				& \includegraphics[width=10em, valign=m]{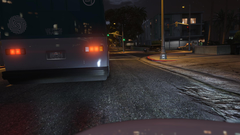}
				& \includegraphics[width=10em, valign=m]{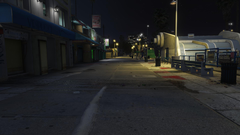}
				& \includegraphics[width=10em, valign=m]{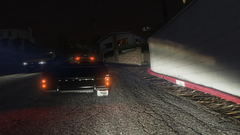}
				\\
				&\adjustbox{valign=m}{\rotatebox{90}{Source}}
				& \includegraphics[width=10em, valign=m]{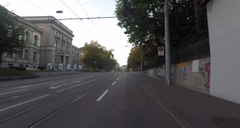}
				& \includegraphics[width=10em, valign=m]{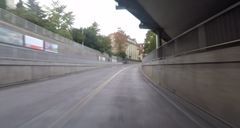}
				& \includegraphics[width=10em, valign=m]{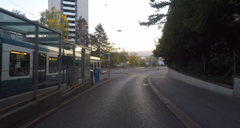}
				\\
				\multicolumn{5}{c}{\textbf{General}}\\\vspace{-3px}
				\adjustbox{valign=m}{\rotatebox{90}{~\cite{liu2019few}}}
				& \adjustbox{valign=m}{\rotatebox{90}{FUNIT}}
				& \includegraphics[width=10em, valign=m]{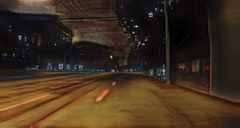}
				& \includegraphics[width=10em, valign=m]{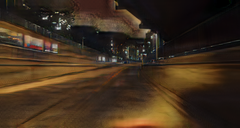}
				& \includegraphics[width=10em, valign=m]{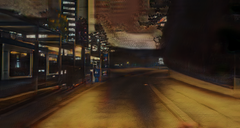}
				\\\vspace{-3px}
				\adjustbox{valign=m}{\rotatebox{90}{~\cite{saito2020coco}}}
				&\adjustbox{valign=m}{\rotatebox{90}{\tiny{COCOFUNIT}}}
				& \includegraphics[width=10em, valign=m]{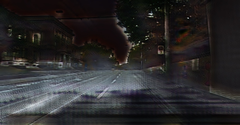}
				& \includegraphics[width=10em, valign=m]{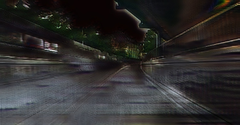}
				& \includegraphics[width=10em, valign=m]{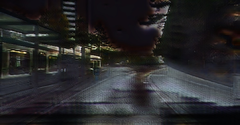}
				\\
				&\adjustbox{valign=m}{\rotatebox{90}{Ours}}
				& \includegraphics[width=10em, valign=m]{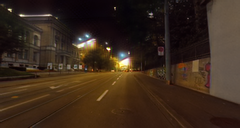}
				& \includegraphics[width=10em, valign=m]{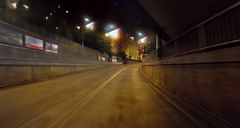}
				& \includegraphics[width=10em, valign=m]{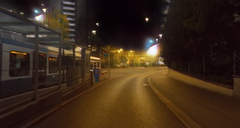}
				\\
				\multicolumn{5}{c}{\textbf{Exemplar}}\\\vspace{-3px}
				\adjustbox{valign=m}{\rotatebox{90}{~\cite{liu2019few}}}
				&\adjustbox{valign=m}{\rotatebox{90}{FUNIT}}
				& \includegraphics[width=10em, valign=m]{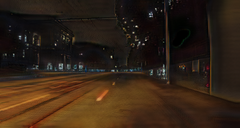}
				& \includegraphics[width=10em, valign=m]{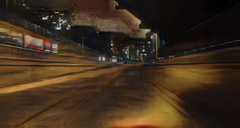}
				& \includegraphics[width=10em, valign=m]{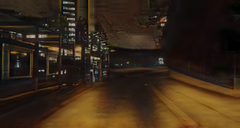}\\\vspace{-3px}
				\adjustbox{valign=m}{\rotatebox{90}{~\cite{saito2020coco}}}
				&\adjustbox{valign=m}{\rotatebox{90}{\tiny{COCOFUNIT}}}
				& \includegraphics[width=10em, valign=m]{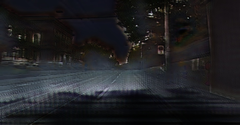}
				& \includegraphics[width=10em, valign=m]{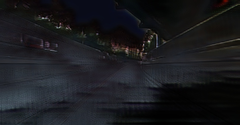}
				& \includegraphics[width=10em, valign=m]{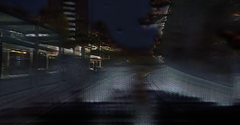}\\

				\adjustbox{valign=m}{\rotatebox{90}{~\cite{yoo2019photorealistic}}}
				&\adjustbox{valign=m}{\rotatebox{90}{$\text{WCT}^2$}}
				& \includegraphics[width=10em, valign=m]{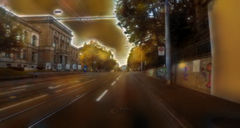}
				& \includegraphics[width=10em, valign=m]{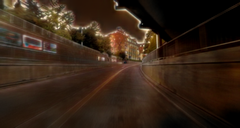}
				& \includegraphics[width=10em, valign=m]{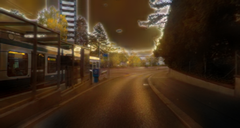}
				\\
				\adjustbox{valign=m}{\rotatebox{90}{~\cite{ma2018exemplar}}}
				&\adjustbox{valign=m}{\rotatebox{90}{EGSC-IT}}
				& \includegraphics[width=10em, valign=m]{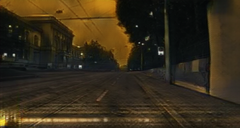}
				& \includegraphics[width=10em, valign=m]{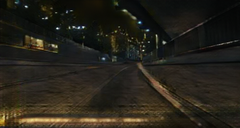}
				& \includegraphics[width=10em, valign=m]{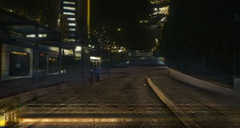}
				\\
				&\adjustbox{valign=m}{\rotatebox{90}{Ours}}
				& \includegraphics[width=10em, valign=m]{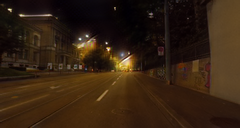}
				& \includegraphics[width=10em, valign=m]{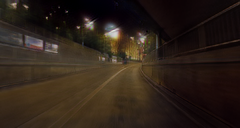}
				& \includegraphics[width=10em, valign=m]{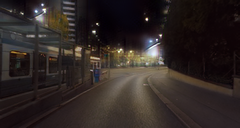}
				\\\midrule
				\adjustbox{valign=m}{\rotatebox{90}{Exemplar}}
				& \adjustbox{valign=m}{\rotatebox{90}{image}}
				& \includegraphics[width=10em, valign=m]{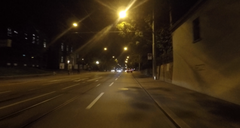}
				& \includegraphics[width=10em, valign=m]{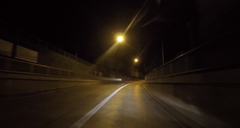}
				& \includegraphics[width=10em, valign=m]{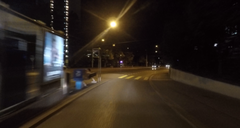}
				\\\bottomrule
				
		\end{tabular}}
		\caption{$\text{Day}\mapsto\text{Night}$}\label{fig:qual-dn}
	\end{subfigure}
	\begin{subfigure}{0.368\linewidth}
		\centering
		\resizebox{\linewidth}{!}{
			\setlength{\tabcolsep}{0.003\linewidth}
			\tiny
			\begin{tabular}{c c c c c}
				\toprule                	&\adjustbox{valign=m}{\rotatebox{90}{Source}}
				& \includegraphics[width=8em, valign=m]{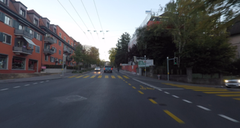}
				& \includegraphics[width=8em, valign=m]{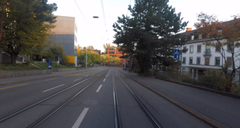}
				& \includegraphics[width=8em, valign=m]{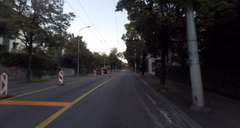}\\
				\multicolumn{5}{c}{\textbf{General}}\\
				\adjustbox{valign=m}{\rotatebox{90}{~\cite{liu2019few}}}
				&\adjustbox{valign=m}{\rotatebox{90}{FUNIT}}
				& \includegraphics[width=8em, valign=m]{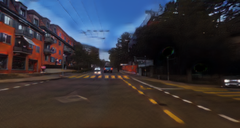}
				& \includegraphics[width=8em, valign=m]{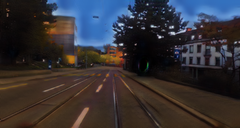}
				& \includegraphics[width=8em, valign=m]{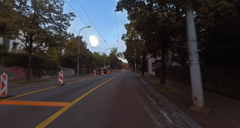}
				\\
				&\adjustbox{valign=m}{\rotatebox{90}{Ours}}   
				& \includegraphics[width=8em, valign=m]{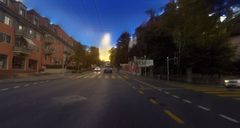}
				& \includegraphics[width=8em, valign=m]{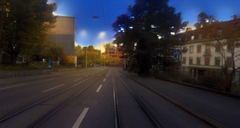}
				& \includegraphics[width=8em, valign=m]{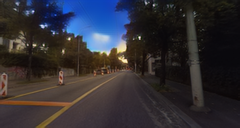}\\
				\multicolumn{5}{c}{\textbf{Exemplar}}\\
				\adjustbox{valign=m}{\rotatebox{90}{~\cite{li2020few}}}
				& \adjustbox{valign=m}{\rotatebox{90}{FUNIT}}
				& \includegraphics[width=8em, valign=m]{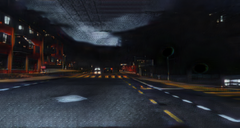}
				& \includegraphics[width=8em, valign=m]{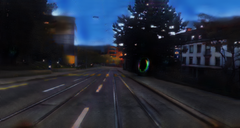}
				& \includegraphics[width=8em, valign=m]{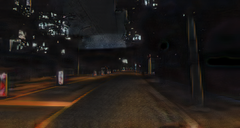}
				\\
				\adjustbox{valign=m}{\rotatebox{90}{~\cite{yoo2019photorealistic}}}
				& \adjustbox{valign=m}{\rotatebox{90}{$\text{WCT}^2$}}
				& \includegraphics[width=8em, valign=m]{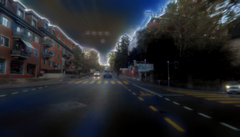}
				& \includegraphics[width=8em, valign=m]{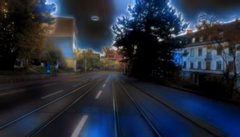}
				& \includegraphics[width=8em, valign=m]{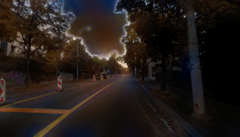}
				\\
				& \adjustbox{valign=m}{\rotatebox{90}{Ours}}
				& \includegraphics[width=8em, valign=m]{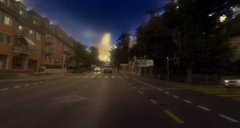}
				& \includegraphics[width=8em, valign=m]{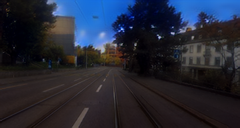}
				& \includegraphics[width=8em, valign=m]{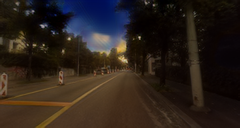}\\\midrule
				\adjustbox{valign=m}{\rotatebox{90}{Exemplar}}
				& \adjustbox{valign=m}{\rotatebox{90}{image}}
				& \includegraphics[width=8em, valign=m]{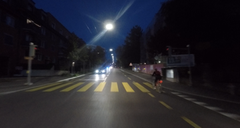}
				& \includegraphics[width=8em, valign=m]{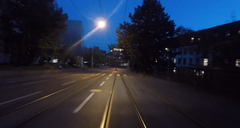}
				& \includegraphics[width=8em, valign=m]{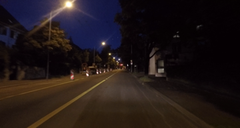}\\
				\bottomrule
		\end{tabular}}
		\caption{$\text{Day}\mapsto\text{Twilight}$}\label{fig:qual-dt}
		\vspace{0.5em}
		\resizebox{\linewidth}{!}{
			\setlength{\tabcolsep}{0.003\linewidth}
			\tiny
			\begin{tabular}{c c c c c}
				\toprule
				& \adjustbox{valign=m}{\rotatebox{90}{Source}}
				& \includegraphics[width=8em, valign=m]{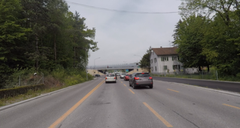}
				& \includegraphics[width=8em, valign=m]{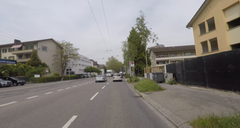}
				& \includegraphics[width=8em, valign=m]{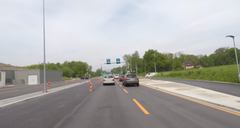}\\
				\multicolumn{5}{c}{\textbf{General}}\\
				\adjustbox{valign=m}{\rotatebox{90}{~\cite{liu2019few}}}
				& \adjustbox{valign=m}{\rotatebox{90}{FUNIT}}
				& \includegraphics[width=8em, valign=m]{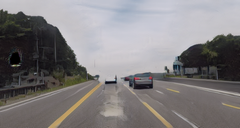}
				& \includegraphics[width=8em, valign=m]{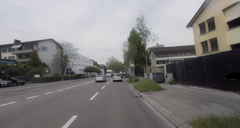}
				& \includegraphics[width=8em, valign=m]{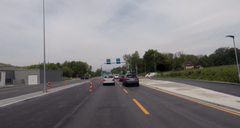}
				\\
				
				& \adjustbox{valign=m}{\rotatebox{90}{Ours}}
				& \includegraphics[width=8em, valign=m]{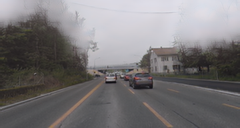}
				& \includegraphics[width=8em, valign=m]{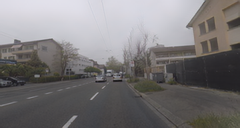}
				& \includegraphics[width=8em, valign=m]{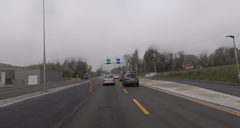}\\
				\multicolumn{5}{c}{\textbf{Exemplar}}\\
				\adjustbox{valign=m}{\rotatebox{90}{~\cite{liu2019few}}}
				& \adjustbox{valign=m}{\rotatebox{90}{FUNIT}}
				& \includegraphics[width=8em, valign=m]{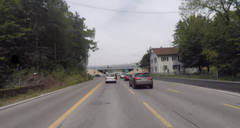}
				& \includegraphics[width=8em, valign=m]{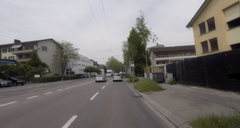}
				& \includegraphics[width=8em, valign=m]{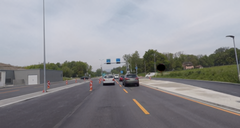}\\
				\adjustbox{valign=m}{\rotatebox{90}{~\cite{yoo2019photorealistic}}}
				& \adjustbox{valign=m}{\rotatebox{90}{$\text{WCT}^2$}}
				& \includegraphics[width=8em, valign=m]{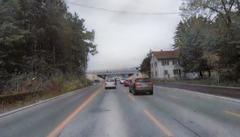}
				& \includegraphics[width=8em, valign=m]{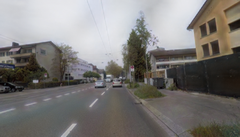}
				& \includegraphics[width=8em, valign=m]{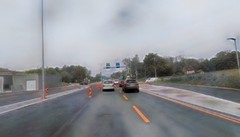}\\
				& \adjustbox{valign=m}{\rotatebox{90}{Ours}}
				& \includegraphics[width=8em, valign=m]{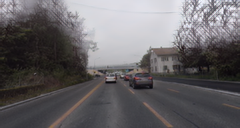}
				& \includegraphics[width=8em, valign=m]{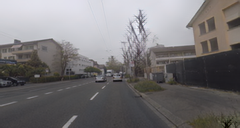}
				& \includegraphics[width=8em, valign=m]{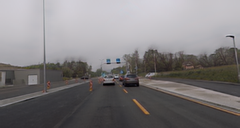}\\\midrule
				\adjustbox{valign=m}{\rotatebox{90}{Exemplar}}
				& \adjustbox{valign=m}{\rotatebox{90}{image}}
				& \includegraphics[width=8em, valign=m]{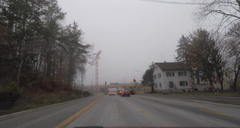}
				& \includegraphics[width=8em, valign=m]{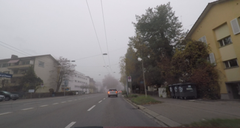}
				& \includegraphics[width=8em, valign=m]{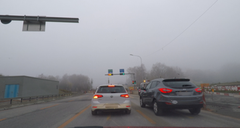}\\
				\bottomrule
		\end{tabular}}
		\caption{$\text{Clear}\mapsto\text{Fog}$}\label{fig:qual-cf}
	\end{subfigure}
	\caption{Qualitative evaluation and comparison with the state of the art. We evaluate the \subref{fig:qual-dn} $\text{Day}\mapsto\text{Night}$, \subref{fig:qual-dt} $\text{Day}\mapsto\text{Twilight}$, and \subref{fig:qual-cf} $\text{Clear}\mapsto\text{Fog}$ tasks. In all cases, our approach extracts a \textit{general} realistic representation of the few-shot target, and correctly reproduces the style of paired \textit{exemplar} target images. In comparison, existing baselines either has unnecessary similarity with anchors (e.g. FUNIT, COCO-FUNIT, EGSC-IT) or unrealistic artifacts (e.g. $\text{WCT}^2$).}
	\label{fig:qual}
\end{figure}

\subsection{Comparison with the state-of-the-art}\label{sec:sota-comparison}

\noindent \textbf{Baselines} \quad
We compare with four baselines for few-shot image translation with $|\mathcal{T}|=25$. We extensively evaluate on the most challenging $\text{Day}\mapsto\text{Night}$ task, and provide insights and comparison for the two others tasks. We evaluate the impact of the few-shot image selection and of $|\mathcal{T}|$ in Sec.~\ref{sec:ablation-images}. We compare against the recent FUNIT~\cite{liu2019few} and COCO-FUNIT~\cite{saito2020coco}, trained on $\mathcal{S}\mapsto\mathcal{A}_m$ and adapted following~\cite{liu2019few,saito2020coco} to the few-shot $\mathcal{T}$ (general) or to a single reference image (exemplar). For exemplar image translation, we also add specific baselines. First, we compare with $\text{WCT}^2$~\cite{yoo2019photorealistic}, used to transfer the style of the paired target condition to the source one. 
We also evaluate EGSC-IT~\cite{ma2018exemplar}. The method is trained by merging $\mathcal{A}_m$ and $\mathcal{T}$ since it should be able to identify inter-domain variability, separating $\mathcal{T}$ styles from $\mathcal{A}_m$~\cite{ma2018exemplar}.
To define metrics bounds, we also train our MUNIT~\cite{huang2018multimodal} backbone on $\mathcal{A}_m$, on the full $\mathcal{T}$ set and on $\mathcal{T}$ with $|\mathcal{T}|=25$. More comparisons with the backbone are in Sec.~\ref{sec:ablation}.
We use the official code provided by the authors for all\footnote{For FUNIT~\cite{liu2019few} and COCO-FUNIT~\cite{saito2020coco}, we modify hyperparameters per authors suggestions to adapt to the ACDC and Dark Zurich datasets.}. More details on baselines are in supp.

\noindent \textbf{Evaluation \:} We compare qualitative results in Fig.~\ref{fig:qual}. In $\text{Day}\mapsto\text{Night}$ (Fig.~\ref{fig:qual-dn}), even if the appearance of images in $\mathcal{T}$ is partially transferred on translated images (e.g. road color, darker sky), FUNIT and COCO-FUNIT still retain some characteristics of $\mathbb{A}$ (note, for example, how the street is similar to the GTA one) which worsens the overall image realism. 
\begin{table}[t!]
\centering
		\begin{subfigure}{0.49\linewidth}
			\centering
		    \resizebox{\linewidth}{!}{%
				\centering
				\scriptsize
				\begin{tabular}{cccccc}
					\toprule
					&\textbf{Method} & \textbf{$|\mathcal{A}_m|$} & \textbf{$|\mathcal{T}|$} & \textbf{FID$_\downarrow$} & \textbf{LPIPS$_\downarrow$} \\\midrule
					\multirow{6}{*}{\textbf{G}} & MUNIT~\cite{huang2018multimodal}
					& 0 & 400 & 79.20 & 0.529 \\
					& MUNIT~\cite{huang2018multimodal}
					& 3090 & 0 & 132.72 & 0.613\\\cmidrule{2-6}
					& MUNIT~\cite{huang2018multimodal}
					& 0 & 25 & 91.61 & 0.553 \\
					& FUNIT~\cite{liu2019few}
					& 3090 & 25 & 156.97 & 0.573 \\
					& COCO-FUNIT~\cite{saito2020coco}
					& 3090 & 25 & 201.67 & 0.644 \\
					& Ours & 3090 & 25 & \textbf{81.01} & \textbf{0.535} \\  %
					\midrule
					\multirow{6}{*}{\vspace{-3.6em}\textbf{E}} & MUNIT~\cite{huang2018multimodal}
					& 0 & 400 & 87.71 & 0.522 \\
					& MUNIT~\cite{huang2018multimodal}
					& 3090 & 0 & 142.04 & 0.559\\\cmidrule{2-6}
					& MUNIT~\cite{huang2018multimodal}
					& 0 & 25 & 128.73 & 0.562 \\
					& FUNIT~\cite{liu2019few}
					& 3090 & 25 & 136.2 & 0.572\\
					& COCO-FUNIT~\cite{saito2020coco}
					& 3090 & 25 & 193.4 & 0.646\\
					& EGSC-IT~\cite{ma2018exemplar}
					& 3090 & 25 & 106.68 & 0.574\\
					& $\text{WCT}^2$~\cite{yoo2019photorealistic}
					& - & - & 105.58 & 0.580\\
					& Ours & 3090 & 25 & \textbf{80.57} & \textbf{0.525} \\\bottomrule  %
					
				\end{tabular}
	            }%
			\caption{$\text{Day}\mapsto\text{Night}$}\label{tab:day2night-quant}
		\end{subfigure} %
		\begin{subfigure}{0.49\linewidth}
			\centering
		    \resizebox{0.8\linewidth}{!}{%

			\scriptsize
			\begin{tabular}{c>{\centering\arraybackslash}p{1.5cm}cc>{\centering\arraybackslash}p{0.7cm}c}
				\toprule
				&\textbf{Method} & \textbf{$|\mathcal{A}_m|$} & \textbf{$|\mathcal{T}|$} & \textbf{FID$_\downarrow$} & \textbf{LPIPS$_\downarrow$} \\\midrule
				\multirow{2}{*}{\textbf{G}} & FUNIT~\cite{liu2019few} & 3090 & 25 &  69.53 & 0.511\\
				&Ours & 3090 & 25 & \textbf{63.15} & \textbf{0.510} \\\midrule
				\multirow{3}{*}{\textbf{E}} & FUNIT~\cite{liu2019few} & 3090 & 25 & 69.97 & 0.501\\ & $\text{WCT}^2$~\cite{yoo2019photorealistic} & - & -  &  71.77 &  0.536\\
				&Ours & 3090 & 25 & \textbf{58.07} & \textbf{0.483} \\\bottomrule  %
			\end{tabular}
	        }%
		\caption{$\text{Day}\mapsto\text{Twilight}$}\label{tab:day2twilight-quant}\vspace{-5px}
			\centering
		    \resizebox{0.8\linewidth}{!}{%

			\scriptsize
			\begin{tabular}{c>{\centering\arraybackslash}p{1.5cm}cc>{\centering\arraybackslash}p{0.7cm}c}
				\toprule
				&\textbf{Method} & \textbf{$|\mathcal{A}_m|$} & \textbf{$|\mathcal{T}|$} & \textbf{FID$_\downarrow$} & \textbf{LPIPS$_\downarrow$} \\\midrule
				\multirow{2}{*}{\textbf{G}} & FUNIT~\cite{liu2019few} & 3090 & 25 & 152.90 & 0.580\\
				&Ours & 3090 & 25 & \textbf{89.57} & \textbf{0.520} \\\midrule
				\multirow{3}{*}{\textbf{E}} & FUNIT~\cite{liu2019few} & 3090 & 25 & 137.7 & 0.568\\
				& $\text{WCT}^2$~\cite{yoo2019photorealistic} & - & - & 120.9 & 0.591 \\
				&Ours & 3090 & 25 & \textbf{89.89} & \textbf{0.521} \\\bottomrule  %
			\end{tabular}
			}%
			\caption{$\text{Clear}\mapsto\text{Fog}$}\label{tab:clear2fog-quant}
		\end{subfigure}
		\caption{Quantitative comparison with state of the art. We compare FID and LPIPS on the (\subref{tab:day2night-quant}) $\text{Day}\mapsto\text{Night}$, (\subref{tab:day2twilight-quant}) $\text{Day}\mapsto\text{Twilight}$ and (\subref{tab:clear2fog-quant}) $\text{Clear}\mapsto\text{Fog}$ tasks, for both \textbf{G}{eneral} and \textbf{E}{xemplar} translations. Our approach outperforms all baselines on all tasks, while also being on par (\textbf{G}) or even outperforming (\textbf{E}) the MUNIT backbone trained on the full dataset for $\text{Day}\mapsto\text{Night}$ in \subref{tab:day2night-quant}.}
		\label{tab:sota-comparison}
	\end{table}%

The same can be observed with EGSC-IT, where the hood of the ego-vehicle in anchor images (first column) is retained and significantly impacts visual results. While $\text{WCT}^2$ exhibits sharp results, it does not correctly map the image context, and it is limited to appearance alignment which leads to artifacts (e.g. yellow sky with white halos). 
Our method generates significantly better results than the baselines in both the \textit{general} and \textit{exemplar} modalities, with visible differences in all three tasks: the \textit{general} appearance is consistent across test samples, and each result adapts to its \textit{exemplar}. For example, observe how the overall sky colors ($\text{Day}\mapsto\text{Twilight}$, Fig.~\ref{fig:qual-dt}) match the exemplar. Here, the exemplars were unseen in training (not part of the few-shot set $\mathcal{T}$), thus GERM generalizes the few-shot learned exemplar behavior. The quantitative evaluation in Tab.~\ref{tab:sota-comparison} is coherent with the qualitative results, as we always outperform baselines. We perform on par (\textit{general}), or even better (\textit{exemplar}) than the backbone trained on the entire set of 400 training images on $\text{Day}\mapsto\text{Night}$ (Tab.~\ref{tab:day2night-quant}).
This result shows that GERM (Sec.~\ref{sec:germ}) improves modeling of the exemplar style over AdaIN exemplar style injection~\cite{huang2018multimodal}. The exemplar behavior may force artifacts following subtle characteristics of the scene (as trees in Fig.~\ref{fig:qual-cf}), for which the general translation may be advisable.

	\begin{figure}[t!]
		\centering
		\begin{subfigure}{0.34\linewidth}
			
			\resizebox{\linewidth}{!}{%
				
				\centering
				\scriptsize
				\setlength{\tabcolsep}{0.017\linewidth}
				\begin{tabular}{ccc}
					\toprule
					\textbf{Model}  & \textbf{mIoU \% $\uparrow$} & \textbf{Acc. \% $\uparrow$} \\\midrule
					Baseline \tiny{(\textit{CS day})} & 12.93 & 45.15\\\midrule
					MUNIT~\cite{huang2018multimodal} & 21.22 & 56.65\\
					Ours & \textbf{24.31} & \textbf{60.50}\\\midrule
					Oracle \tiny{(\textit{ACDC night})} & 49.23 & 88.47\\
					\bottomrule
				\end{tabular}
			}%
			\caption{Quantitative evaluation}\label{fig:segmentation-quant}
		\end{subfigure}\hfill%
		\begin{subfigure}{0.65\linewidth}
			\centering
			\resizebox{1\linewidth}{!}{
				\setlength{\tabcolsep}{0.002\linewidth}
				\scriptsize
				\begin{tabular}{cccc}
					Input & GT & MUNIT & Ours \\
					\includegraphics[width=8em, valign=m,cframe=green 0.4mm]{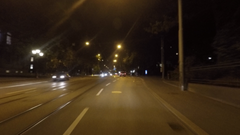}&\includegraphics[width=8em, valign=m,cframe=green 0.4mm]{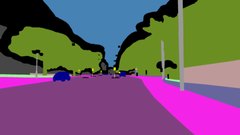} %
					&\includegraphics[width=8em, valign=m]{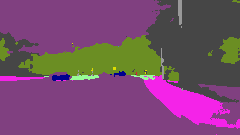}  %
					&\includegraphics[width=8em, valign=m]{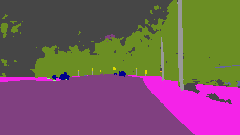}%
				\end{tabular}
			}%
			\caption{Qualitative evaluation}\label{fig:segmentation-qual}
		\end{subfigure}
		\caption{Segmentation on ACDC-night, for few-shot $\text{Day}{\mapsto}\text{Night}$ translations (${|\mathcal{T}| = 25}$) (\subref{tab:ablation-quant}). We outperform the baseline with noticeably better segmentation in (\subref{tab:ablation-qual}) due to the increased quality of our translation.} %
	\label{fig:segmentation}
\end{figure}
\subsection{Segmentation downstream task}
\label{sec:exp-seg}
\noindent We exploit semantic segmentation to evaluate \name for increasing robustness in challenging scenarios. 
In Fig.~\ref{fig:segmentation} we train HRNet~\cite{WangSCJDZLMTWLX19} on nighttime versions of Cityscapes~\cite{cordts2016cityscapes} obtained by translating the dataset with \name or MUNIT, and evaluating on the ACDC-night validation set labels. We choose the best MUNIT and \name configurations following nighttime realism in Tab.~\ref{tab:day2night-quant} with $|\mathcal{T}|=25$. As lower and upper bounds we train HRNet either on original Cityscapes (baseline) or on ACDC-night training set (oracle). Fig.~\ref{fig:segmentation} shows we outperform the MUNIT backbone (+3.09 mIoU) thanks to our better target domain modeling. Additional results on other domains are in supplementary.

\begin{figure}[t]
	\centering
	\setlength{\tabcolsep}{0.003\linewidth}
	\tiny
	\begin{subfigure}{0.49\linewidth}
		\resizebox{\linewidth}{!}{%
			\begin{tabular}{c c c c c | c c}
				\adjustbox{valign=m}{\rotatebox{90}{Source}}
				& \includegraphics[width=5em, valign=m,cframe=green 0.4mm]{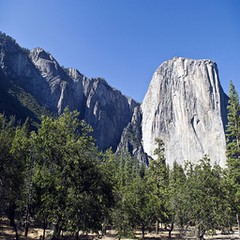}
				& \includegraphics[width=5em, valign=m]{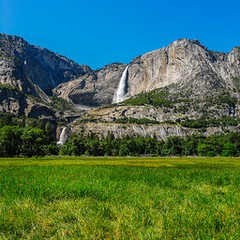}
				& \includegraphics[width=5em, valign=m]{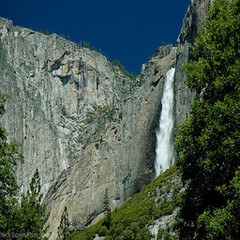}
				& \includegraphics[width=5em, valign=m]{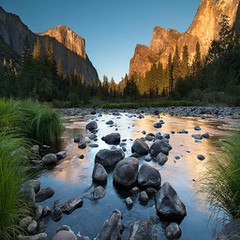}
				& \multicolumn{2}{c}{\includegraphics[width=5em, height=5em, valign=m]{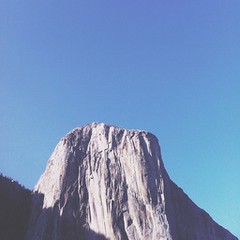}}
				\\[2.5em]
				\adjustbox{valign=m}{\rotatebox{90}{FUNIT~\cite{liu2019few}}}
				& \includegraphics[width=5em, valign=m,cframe=green 0.4mm]{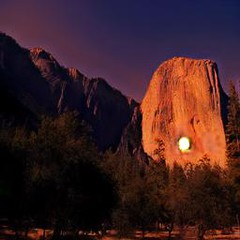}
				& \includegraphics[width=5em, valign=m]{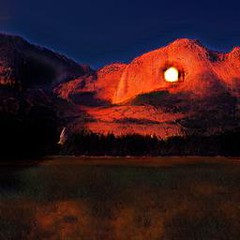}
				& \includegraphics[width=5em, valign=m]{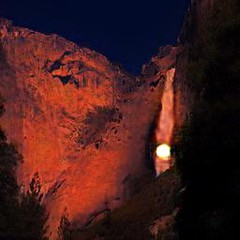}
				& \includegraphics[width=5em, valign=m]{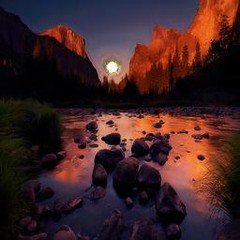}
				& \includegraphics[width=5em, height=5em, valign=m]{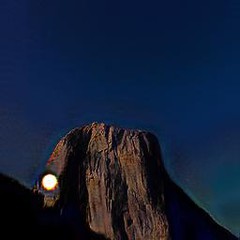}
				& \includegraphics[width=5em, height=5em, valign=m]{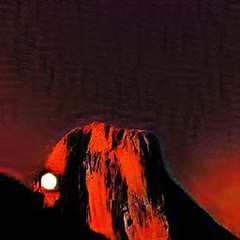}
				\\
				\adjustbox{valign=m}{\rotatebox{90}{Ours}}
				& \includegraphics[width=5em, valign=m,cframe=green 0.4mm]{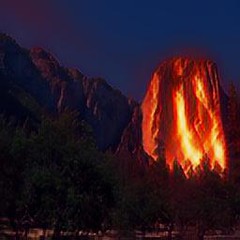}
				& \includegraphics[width=5em, valign=m]{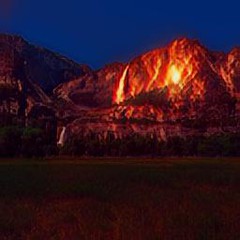}
				& \includegraphics[width=5em, valign=m]{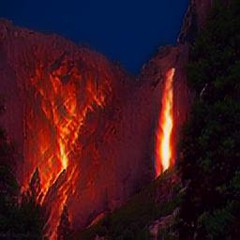}
				& \includegraphics[width=5em, valign=m]{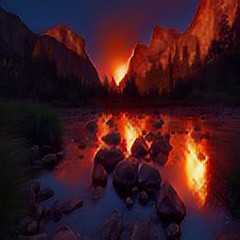}
				& \includegraphics[width=5em, height=5em, valign=m]{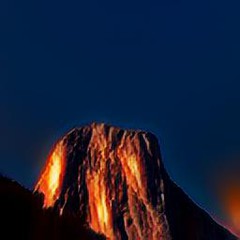}
				& \includegraphics[width=5em, height=5em, valign=m]{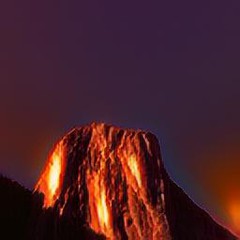}\\[2.5em]
				\adjustbox{valign=m}{\rotatebox{90}{Target}}
				& \includegraphics[width=5em, valign=m,cframe=green 0.4mm]{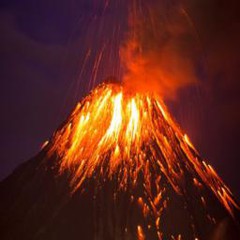}
				& \includegraphics[width=5em, valign=m]{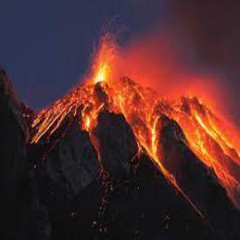}
				& \includegraphics[width=5em, valign=m]{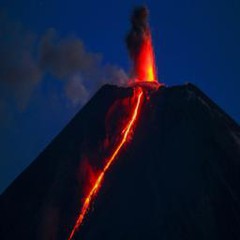}
				& \includegraphics[width=5em, valign=m]{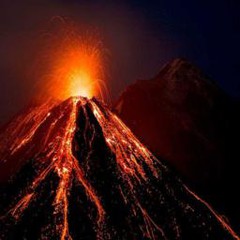}
				& \includegraphics[width=5em, height=5em, valign=m]{figures/figure_volcano/volc_3.jpg}
				& \includegraphics[width=5em, height=5em, valign=m]{figures/figure_volcano/volc_4.jpg}\\
				\cmidrule[1pt](lr){2-5}\cmidrule[1pt](lr){6-7}
				&\multicolumn{4}{c}{General} & \multicolumn{2}{c}{Exemplar}
			\end{tabular}
		} %
		\caption{$\text{Mountain}\mapsto\text{Volcano}$}\label{fig:volcano-subfig1}
	\end{subfigure}
	\hfill
	\begin{subfigure}{0.49\linewidth}
		\resizebox{\linewidth}{!}{%
			
			\begin{tabular}{c c c c c | c c}
				
				\adjustbox{valign=m}{\rotatebox{90}{Source}}
				& \includegraphics[width=5em, valign=m,cframe=green 0.4mm]{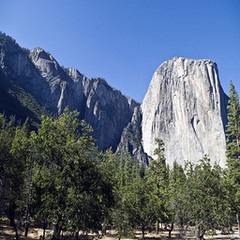}
				& \includegraphics[width=5em, valign=m]{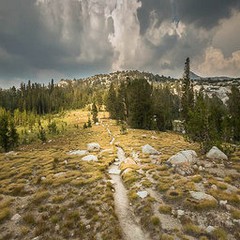}
				& \includegraphics[width=5em, valign=m]{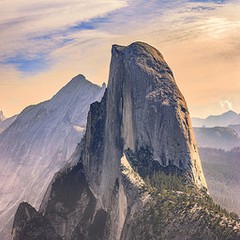}
				& \includegraphics[width=5em, valign=m]{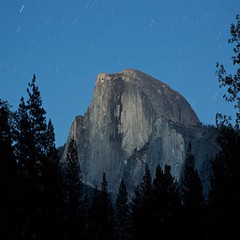}
				& \multicolumn{2}{c}{\includegraphics[width=5em, height=5em, valign=m]{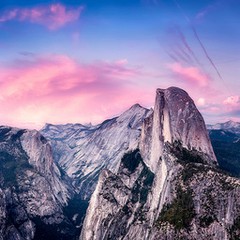}}
				\\[2.5em]
				\adjustbox{valign=m}{\rotatebox{90}{FUNIT~\cite{liu2019few}}}
				& \includegraphics[width=5em, valign=m,cframe=green 0.4mm]{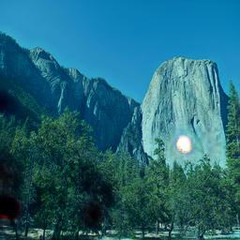}
				& \includegraphics[width=5em, valign=m]{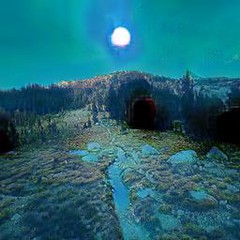}
				& \includegraphics[width=5em, valign=m]{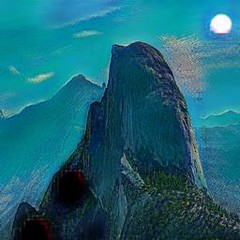}
				& \includegraphics[width=5em, valign=m]{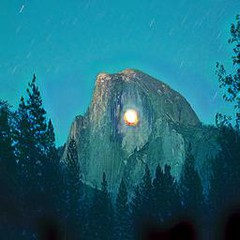}
				& \includegraphics[width=5em, height=5em, valign=m]{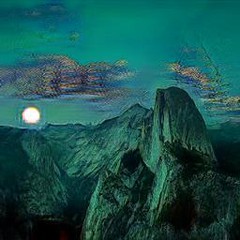}
				& \includegraphics[width=5em, height=5em, valign=m]{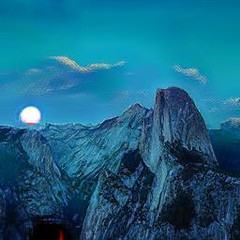}\\
				\adjustbox{valign=m}{\rotatebox{90}{Ours}}
				& \includegraphics[width=5em, valign=m,cframe=green 0.4mm]{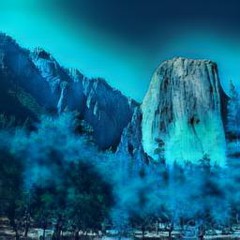}
				& \includegraphics[width=5em, valign=m]{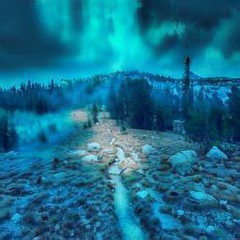}
				& \includegraphics[width=5em, valign=m]{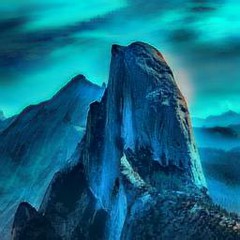}
				& \includegraphics[width=5em, valign=m]{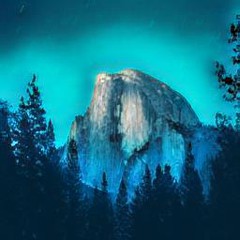}
				& \includegraphics[width=5em, height=5em, valign=m]{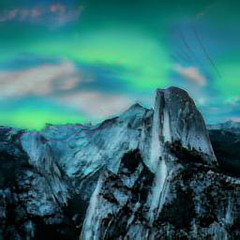}
				& \includegraphics[width=5em, height=5em, valign=m]{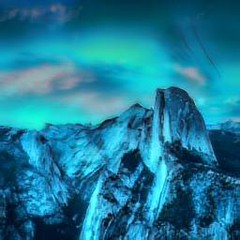}\\[2.5em]
				\adjustbox{valign=m}{\rotatebox{90}{Target}}
				& \includegraphics[width=5em, valign=m,cframe=green 0.4mm]{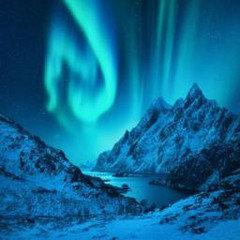}
				& \includegraphics[width=5em, valign=m]{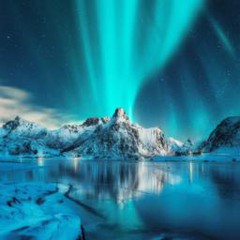}
				& \includegraphics[width=5em, valign=m]{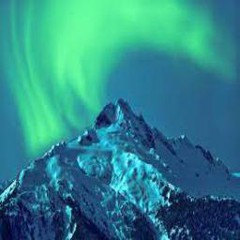}
				& \includegraphics[width=5em, valign=m]{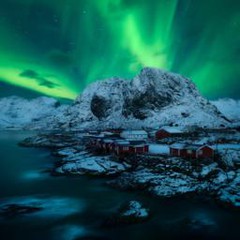}
				& \includegraphics[width=5em, height=5em, valign=m]{figures/figure_northern/northern_4.jpg}
				& \includegraphics[width=5em, height=5em, valign=m]{figures/figure_northern/northern_1.jpg}\\
				
				\cmidrule[1pt](lr){2-5}\cmidrule[1pt](lr){6-7}
				&\multicolumn{4}{c}{General} & \multicolumn{2}{c}{Exemplar}\\
				
			\end{tabular}
		}%
		\caption{$\text{Day}\mapsto\text{Aurora}$}\label{fig:volcano-subfig2}
	\end{subfigure}
	
	\caption{Qualitative results for the $\text{Mountain}\mapsto\text{Volcano}$ (\subref{fig:volcano-subfig1}) and $\text{Day}\mapsto\text{Aurora}$ (\subref{fig:volcano-subfig2}) tasks. We retain contextual information by only partially mapping mountains to volcanoes and sky to auroras. In the green box we process the same image for ease of comparison. \emph{Exemplar} results show how Ours conforms to Target, effectively reproducing the exemplar image style (cols 5--6). For space reason, we show WCT$^2$~\cite{yoo2019photorealistic} outputs in Supp.}\label{fig:volcano}
\end{figure}

\subsection{Rare few-shot scenarios}
\label{sec:exp-volcanoes}
Few-shot plays its full role with conditions that are rare by nature, difficult or even dangerous to photograph, such as auroras or erupting volcanoes. Fig.~\ref{fig:volcano} shows the capability of \name to learn $\text{Mountain}{\mapsto}\text{Volcano}$ or $\text{Day}{\mapsto}\text{Aurora}$, by taking as source and anchor the summer and winter Yosemite dataset~\cite{zhu2017unpaired} splits respectively. Each task uses only 4 images from Google Images as $\mathcal{T}$. 
We generate realistic erupting volcanoes or auroras starting from mountain images, with contextual understanding (Fig.~\ref{fig:volcano}, cols 1--4), where only one mountain is mapped to a volcano and auroras only partially cover the sky. Fig.~\ref{fig:volcano} (cols 5--6) also demonstrate how exemplar characteristics are preserved.
\subsection{Ablation studies}\label{sec:ablation}

\begin{figure}[t]
	\centering
	\begin{subfigure}{0.45\linewidth}

			\centering
			\scriptsize
			\setlength{\tabcolsep}{0.017\linewidth}
			\begin{tabular}{ccc}
				\toprule
				\textbf{Component}  & \textbf{FID$_\downarrow$} & \textbf{LPIPS$_\downarrow$} \\\midrule
				w/o $\mathcal{L}_{\text{style}}$ & 143.66 & 0.614\\
				w/o $\mathcal{L}_{\text{patch}}$ & 93.42 & 0.566 \\
				w/o GERM & 85.62 & 0.544\\\midrule
				w/o WMI & 101.57 & 0.589\\
				LGFS-only & 84.29 & 0.558\\\midrule
				Ours & \textbf{81.01} & \textbf{0.535}\\        
				\bottomrule
			\end{tabular}
		\caption{Quantitative evaluation}\label{tab:ablation-quant}
	\end{subfigure}\hfill%
	\begin{subfigure}{0.55\linewidth}
		\centering
		\resizebox{1\linewidth}{!}{
			\setlength{\tabcolsep}{0.002\linewidth}
			\tiny
			\begin{tabular}{ccc}
				w/o $\mathcal{L}_{\text{style}}$ & w/o $\mathcal{L}_{\text{patch}}$ & w/o GERM\\
				\includegraphics[width=9em, valign=m]{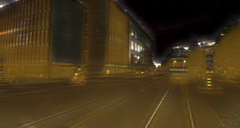}&\includegraphics[width=9em, valign=m]{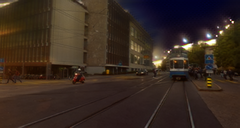}& 
				\includegraphics[width=9em, valign=m]{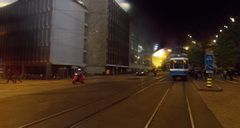}\\
				w/o WMI & LGFS-only & Ours \\
				\includegraphics[width=9em, valign=m]{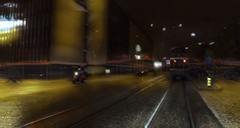}&  
				\includegraphics[width=9em, valign=m]{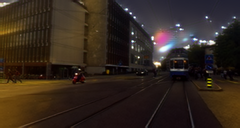}&\includegraphics[width=9em, valign=m]{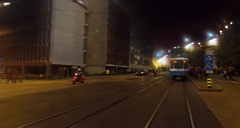}%
			\end{tabular}
		}%
		\caption{Qualitative evaluation}\label{tab:ablation-qual}
	\end{subfigure}
	\caption{Ablation study for architectural components. \subref{tab:ablation-quant} Removing each component individually lowers quantitative performances, which maps to \subref{tab:ablation-qual} decreased visual quality in the generated images.} %
\label{tab:ablation}
\end{figure}
\noindent \textbf{Architectural components} \quad
We evaluate the contribution of each component in \name (\cf Fig.~\ref{fig:architecture}, Sec.~\ref{sec:method}) using the $\text{Day}\mapsto\text{Night}$ task in the \textit{general} scenario, and report results in Fig.~\ref{tab:ablation-quant}. 
The impact of LGFS is studied by removing $\mathcal{L}_{\text{style}}$ or $\mathcal{L}_{\text{patch}}$, showing that both local \textit{and} global guidance are improving translations. Removing the GERM from the training pipeline simultaneously precludes the \textit{exemplar} behavior and worsens the performance, demonstrating the effectiveness of encoding complementary characteristics outside of the manifold spanned by $\mathbb{A}$.  
The benefit of WMI is evaluated in two experiments. First, the ``w/o WMI'' setting applies the residual directly on the fake anchor images $\tilde{s}_c$, instead of the interpolated $\tilde{s}_w$ as in Eq.~(\ref{eqn:residual}). The worse performance relate to synthetic characteristics present in $\tilde{s}_c$ (\eg road texture in \cref{tab:ablation-qual}). Second, ``LGFS-only'' directly uses the LGFS losses in substitution to $\mathcal{L}_{\text{adv}}$, without WMI and GERM components. While it only slightly worsens metrics, the impact on feature consistency is dramatic as shown in \cref{tab:ablation-qual}, where the sky presents obvious artifacts and road trivially darkens. 

\begin{table}[t]
\centering
\begin{subfigure}{0.70\linewidth}
	\resizebox{0.31\linewidth}{!}{%
		\begin{tabular}{Hccccc}
    	    \multicolumn{6}{c}{$\text{Day}\mapsto\text{Night}$}\\
			\toprule
			&\textbf{$\mathcal{S}$} & \textbf{$\mathcal{T}$} & \textbf{$\mathcal{A}_m$}  & \textbf{FID$_\downarrow$} & \textbf{LPIPS$_\downarrow$} \\\midrule
			\multirow{6}{*}{\rotatebox{90}{$\text{Day}\mapsto\text{Night}$}} & \multirow{6}{*}{\scriptsize\rotatebox{90}{ACDC-Day}~} & \multirow{6}{*}{\scriptsize\rotatebox{90}{ACDC-Night}~} & Day & 85.73 & 0.553\\
			&&& Night & \textbf{81.01} & \textbf{0.535}\\ %
			&&& Rain & \underline{81.38} & 0.549\\
			&&& Snow & 86.74 & 0.554\\
			&&& Sunset & 83.83 & 0.571\\
			&&& All & 83.71 & \underline{0.547}\\\bottomrule  
			\arrayrulecolor{green}%
			\cmidrule[1.5pt](lr){1-6}
			\arrayrulecolor{black}%

		\end{tabular}

	}%
	\hfill
	\resizebox{0.31\linewidth}{!}{%
		\begin{tabular}{Hccccc}
    	    \multicolumn{6}{c}{$\text{Day}\mapsto\text{Twilight}$}\\
			\toprule
			&\textbf{$\mathcal{S}$} & \textbf{$\mathcal{T}$} & \textbf{$\mathcal{A}_m$}  & \textbf{FID$_\downarrow$} & \textbf{LPIPS$_\downarrow$} \\\midrule
			\multirow{6}{*}{\rotatebox{90}{$\text{Day}\mapsto\text{Twilight}$}} & \multirow{6}{*}{\rotatebox{90}{\scriptsize{DZ-Day}}~} & \multirow{6}{*}{\rotatebox{90}{\scriptsize{DZ-Twilight}}~} & Day & 64.19 & 0.505 \\
			&&& Night & \underline{63.15} & 0.510 \\
			&&& Rain & 65.33 & \underline{0.501} \\
			&&& Snow & 64.09 & 0.513 \\
			&&& Sunset & 63.78 & 0.504 \\
			&&& All & \textbf{60.98} & \textbf{0.469}\\
			\bottomrule
			\arrayrulecolor{green}%
			\cmidrule[1.5pt](lr){1-6}
			\arrayrulecolor{black}%
			
		\end{tabular}
	}%
	\hfill
	\resizebox{0.31\linewidth}{!}{%
		\begin{tabular}{Hccccc}
    	    \multicolumn{6}{c}{$\text{Clear}\mapsto\text{Fog}$}\\
			\toprule
			&\textbf{$\mathcal{S}$} & \textbf{$\mathcal{T}$} & \textbf{$\mathcal{A}_m$}  & \textbf{FID$_\downarrow$} & \textbf{LPIPS$_\downarrow$} \\\midrule
			\multirow{6}{*}{\rotatebox{90}{$\text{Clear}\mapsto\text{Fog}$}}&\multirow{6}{*}{\rotatebox{90}{\scriptsize{ACDC-Clear}}~} &  \multirow{6}{*}{\rotatebox{90}{\scriptsize{ACDC-Fog}}~} & Day & \textbf{89.57} & \textbf{0.520} \\
			&&& Night & 91.79 & \textbf{0.520} \\
			&&& Rain & 93.15 & \underline{0.522} \\
			&&& Snow & 90.28 & 0.524 \\
			&&& Sunset & \underline{90.11} & 0.525 \\
			&&& All & 92.19 & \textbf{0.520}\\\bottomrule
			\arrayrulecolor{green}%
			\cmidrule[1.5pt](lr){1-6}
			\arrayrulecolor{black}%

		\end{tabular}
	}%
	\caption{{Intra-dataset}}
	\label{tab:ablation-intrads}
\end{subfigure}\hfill
\begin{subfigure}{0.225\linewidth}
	\hfill\hfill
	\resizebox{0.9644\linewidth}{!}{%
		\begin{tabular}{Hccccc}
	    \multicolumn{6}{c}{$\text{Day}\mapsto\text{Twilight}$}\\
			\toprule
			&\textbf{$\mathcal{S}$} & \textbf{$\mathcal{T}$} & \textbf{$\mathcal{A}_m$}  & \textbf{FID$_\downarrow$} & \textbf{LPIPS$_\downarrow$} \\\midrule
			&\multirow{6}{*}{\rotatebox{90}{\scriptsize{ACDC-Day}}~} & \multirow{6}{*}{\rotatebox{90}{\scriptsize{DZ-Twilight}}~} & Day & 89.61 & * \\
			&&& Night & 90.48 & *\\
			&&& Rain & \underline{89.47} & * \\
			&&& Snow & 91.49 & * \\
			&&& Sunset & 91.77 & * \\
			&&& All & \textbf{85.15} & *\\\bottomrule
			\arrayrulecolor{red}%
			\cmidrule[1.5pt](lr){1-6}
			\arrayrulecolor{black}%

		\end{tabular}
	}%
	\caption{{Cross-dataset}}\label{tab:tab-ablation-crossds}
\end{subfigure}
	\caption{Study of the impact of anchor domains $\mathbb{A}$ on the $\mathcal{S}\mapsto\mathcal{T}$ translations for \ulgreen{intra-dataset} (\subref{tab:ablation-intrads}) and \ulred{cross-dataset} (\subref{tab:tab-ablation-crossds}) tasks. The stable performance across all tested anchors demonstrates the robustness of our method. For all, we test a multi-anchor setup by using all anchors (``All''). In~(\subref{tab:tab-ablation-crossds}), * means LPIPS cannot be computed due to lack of pairs of matched images (Sec.~\ref{sec:datasets}).}\label{tab:ablation-anchors}
\end{table}

\noindent \textbf{Anchor selection} \quad
We ablate the choice of anchor domain $\mathbb{A}$ by  selecting different conditions from the VIPER dataset, 
namely \{Day, Night, Rain, Snow, Sunset\}. In particular, we experiment on previous \textit{intra-dataset} ($\mathcal{S}$ and $\mathcal{T}$ taken from the same dataset) tasks, as well as on a \textit{cross-dataset} task in which $\mathcal{S} = \text{ACDC-Day}$ and $\mathcal{T} = \text{DZ-Twilight}$. Results in Tab.~\ref{tab:ablation-anchors} show how performance remains relatively stable across most anchors. This may seem counter-intuitive since one could, for example, expect that the ``Rain'' anchor would be a poor choice for the $\text{Day}\mapsto\text{Night}$ task since rainy and night scenes look different. The results instead show that the WMI only encodes consistency in the transformation, and is thus robust to the choice of anchors. We also test a multi-anchor setup (``All'' in Tab.~\ref{tab:ablation-anchors}), where $\mathbb{A}=$\{$\mathcal{A}_\mathrm{id}$, Day, Night, Rain, Snow, Sunset\}. In general, more anchor domains improve performances, ranking either first or second in all cases for at least one metric, due to the additional information available for shaping the manifold in WMI. We hypothesize that multiple anchors helps identifying correspondences between $\mathcal{S}$ and $\mathcal{T}$, benefiting especially the cross-dataset tasks.

\begin{figure}[t]
	\centering
	\begin{subfigure}{0.45\linewidth}
		\resizebox{\linewidth}{!}{%
			\setlength{\tabcolsep}{0.003\linewidth}
			\tiny
			\begin{tabular}{c c c c}
				\multirow{1}{*}[0em]{\textbf{Source}} && MUNIT & Ours\\
				\multirow{4}{*}[-2.5em]{\includegraphics[width=12em, valign=m]{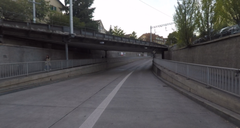}}
				& \adjustbox{valign=m}{\rotatebox{90}{25}}
				& \includegraphics[width=8em, valign=m]{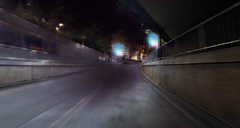}
				& \includegraphics[width=8em, valign=m]{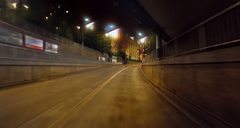}
				\\
				& \adjustbox{valign=m}{\rotatebox{90}{10}} 			& \includegraphics[width=8em, valign=m]{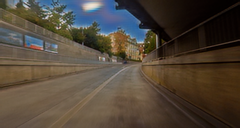}
				& \includegraphics[width=8em, valign=m]{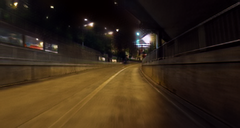}
				\\
				& \adjustbox{valign=m}{\rotatebox{90}{5}} 			& \includegraphics[width=8em, valign=m]{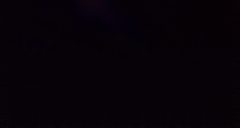}
				& \includegraphics[width=8em, valign=m]{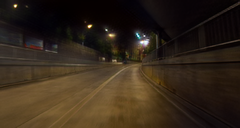}
				\\
				& \adjustbox{valign=m}{\rotatebox{90}{1}} 			& \includegraphics[width=8em, valign=m]{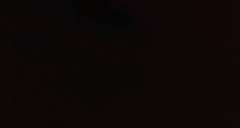}
				& \includegraphics[width=8em, valign=m]{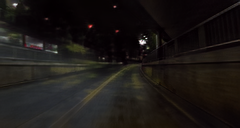}
				\\
				
			\end{tabular}
		}%
		\caption{Qualitative evaluation (general).}\label{fig:munit-comparison-qual}
	\end{subfigure}
	\begin{subfigure}{0.25\linewidth}
		\centering
		\captionsetup{justification=centering}
		\resizebox{\linewidth}{!}{%
			\includegraphics[width=\linewidth]{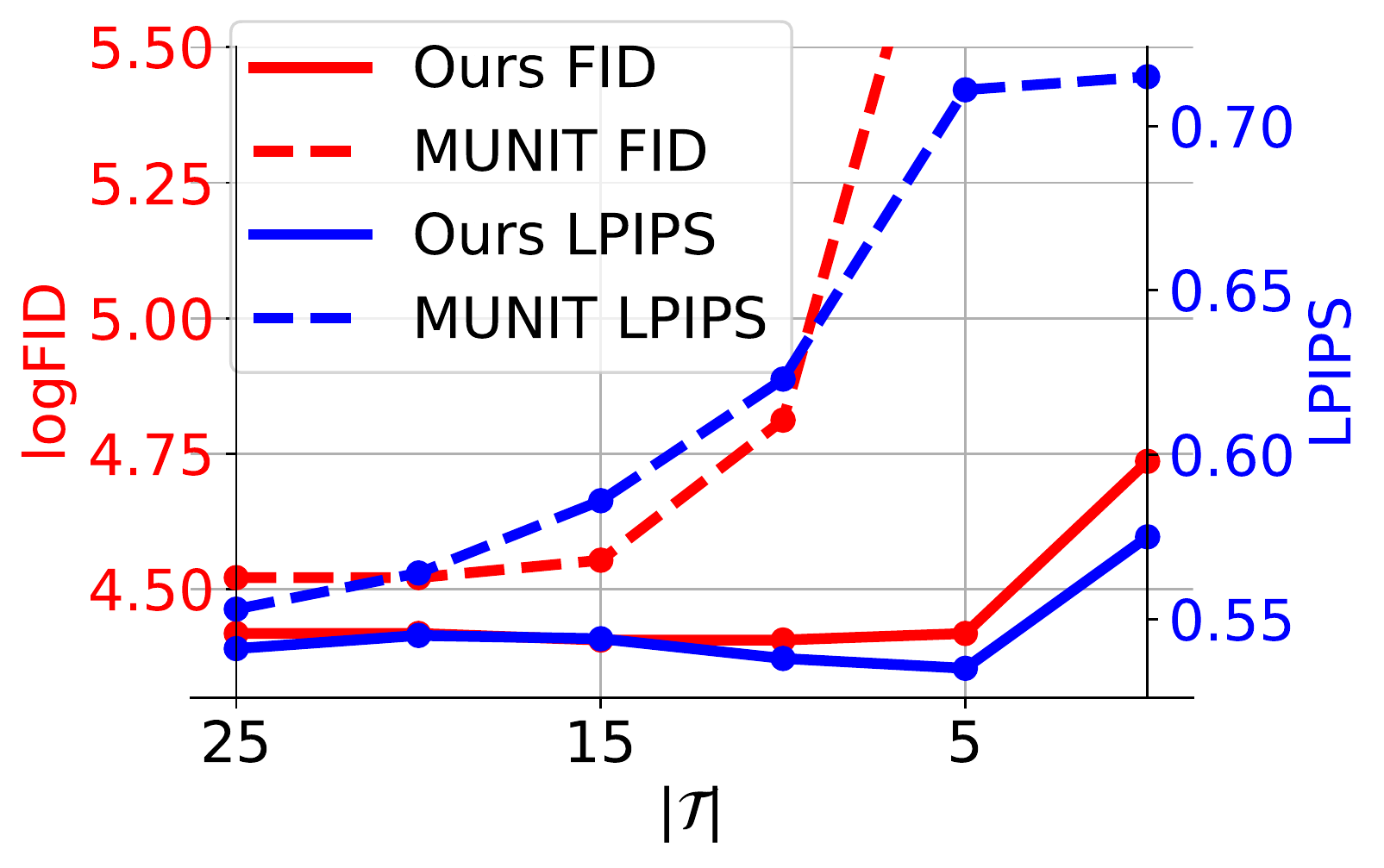}
		}%
		\caption{Quant. evaluation (general)}
		\label{fig:munit_comparison_quant_general}
	\end{subfigure}
	\begin{subfigure}{0.25\linewidth}
		\centering
		\captionsetup{justification=centering}
		\resizebox{\linewidth}{!}{%
			\includegraphics[width=\linewidth]{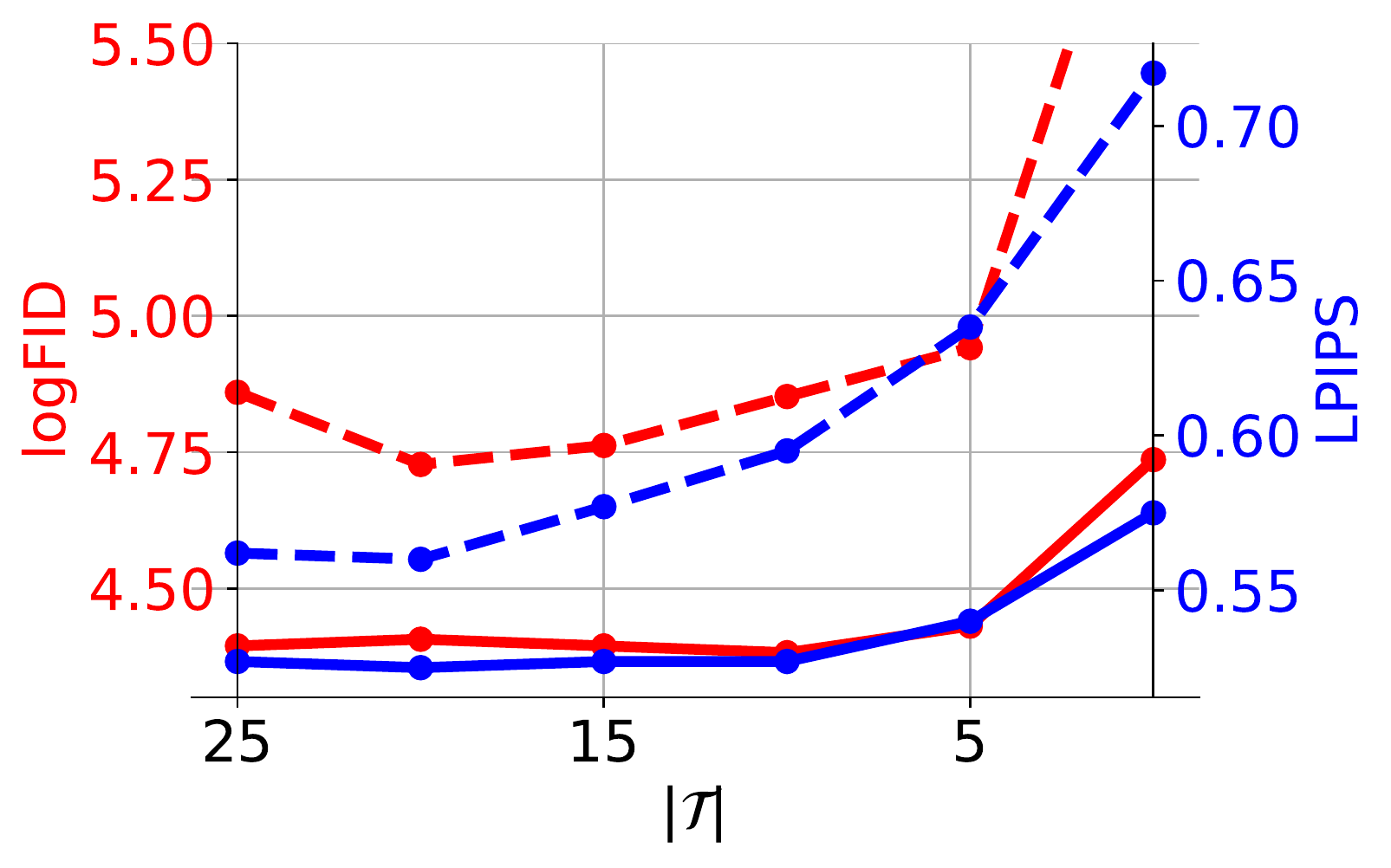}
		}%
		\caption{Quant. evaluation (exemplar)}\label{fig:munit_comparison_quant_exemplar}
	\end{subfigure}
	\caption{Comparison against MUNIT for varying $|\mathcal{T}|$: \subref{fig:munit-comparison-qual} qualitatively for the \textit{general} scenario; as well as quantitatively for \subref{fig:munit_comparison_quant_general} \textit{general} and \subref{fig:munit_comparison_quant_exemplar} \textit{exemplar}, always outperforming it.}
	\label{fig:munit-comparison}
\end{figure}

\noindent \textbf{Number of images and variability} \quad
\label{sec:ablation-images}
First we compare our $\text{Day}\mapsto\text{Night}$ translations against MUNIT~\cite{huang2018multimodal} for \mbox{$|\mathcal{T}|=\{25, 20, 15, 10, 5, 1\}$}, to understand the effects of few-shot training on the backbone network. Some qualitative \textit{general} outputs are shown in Fig.~\ref{fig:munit-comparison-qual}. While MUNIT overfits and creates unrealistic appearance (25--10 images) or collapse (5, 1 image), we output realistic transformations in all cases, even retaining the image context in the extreme one-shot scenario. This is confirmed by the FID and LPIPS in Figs~\ref{fig:munit_comparison_quant_general} and \ref{fig:munit_comparison_quant_exemplar} for the \textit{general} and \textit{exemplar} scenarios respectively.

\begin{table}[t]
	\centering
	\newcommand{\var}[1]{\tiny{$\pm$#1}}
	\begin{subfigure}{0.45\linewidth}

			\centering
			\scriptsize
			\setlength{\tabcolsep}{0.017\linewidth}
			\begin{tabular}{ccc}
				\toprule
				\textbf{$|\mathcal{T}|$} & \textbf{FID$_\downarrow$} & \textbf{LPIPS$_\downarrow$}\\
				\midrule
				25 & 82.95 \var{2.95} & 0.541 \var{1.85e-2} \\  %
				15 & 82.21 \var{3.09} & 0.544 \var{2.35e-2}\\  %
				5 & 83.11 \var{2.49} & 0.535 \var{2.24e-2}\\  %
				1 & 114.5 \var{34.2} & 0.575 \var{2.37e-2}\\  %
				\bottomrule
			\end{tabular}
		\caption{General}\label{tab:tab-ablation-numimages-general}
	\end{subfigure}%
	\begin{subfigure}{0.45\linewidth}

			\centering
			\scriptsize
			\setlength{\tabcolsep}{0.017\linewidth}
			\begin{tabular}{ccc}
				\toprule
				\textbf{$|\mathcal{T}|$} & \textbf{FID$_\downarrow$} & \textbf{LPIPS$_\downarrow$}\\
				\midrule
				25 & 80.78 \var{2.91} & 0.527 \var{0.64e-2} \\  %
				15 & 80.55 \var{2.85} & 0.527 \var{1.07e-2}\\  %
				5 & 84.40 \var{1.88} & 0.540 \var{1.88e-2}\\  %
				1 & 114.3 \var{33.5} & 0.575 \var{2.40e-2}\\  %
				\bottomrule
			\end{tabular}
		\caption{Exemplar}\label{tab:tab-ablation-numimages-exemplar}
	\end{subfigure}\hfill%
	\caption{$\text{Day}\mapsto\text{Night}$ ablation on variability by training on 4 few-shot configurations with 7 runs each on general (\subref{tab:tab-ablation-numimages-general}) and exemplar (\subref{tab:tab-ablation-numimages-exemplar}). $|\mathcal{T}|$ does not impact performance much except for the extreme one-shot scenario, where the network overfits to the seen style. The exemplar behavior performs better due to the style conditioning mechanism.}
	\label{tab:tab-ablation-numimages}
\end{table}

In Tab.~\ref{tab:tab-ablation-numimages} we also study variability, evaluating FID and LPIPS for the \emph{general} and \emph{exemplar} cases for $|\mathcal{T}| = \{25, 15, 5, 1\}$ images reporting the results of 7 runs. 
Overall, the performance remains relatively constant with the exception of the one-shot setup, where despite realistic transfer, the metrics are penalized since the target image itself might not accurately represent the style distribution of the test set.

\begin{figure}[t]
	\centering
	\resizebox{0.7\linewidth}{!}{
		\setlength{\tabcolsep}{0.003\linewidth}
		\tiny
		\begin{tabular}{c c c c c c}
			&& \tikzmark{a}{0} &&& \tikzmark{b}{1}\\
			\multirow{2}{*}{\rotatebox{90}{\textbf{Non few-shot}}}&\adjustbox{valign=m}{\tiny\rotatebox{90}{DNI-M}}
			& \includegraphics[width=10em,height=5em, valign=m]{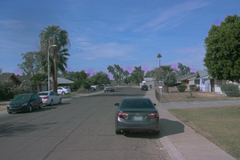}
			& \includegraphics[width=10em,height=5em, valign=m]{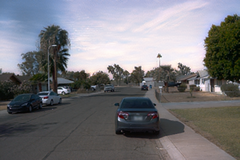}
			& \includegraphics[width=10em,height=5em, valign=m]{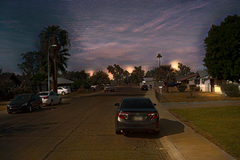}
			& \includegraphics[width=10em,height=5em, valign=m]{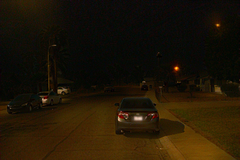}\\
			&\adjustbox{valign=m}{\tiny\rotatebox{90}{CoMoGAN}}
			& \includegraphics[width=10em,height=5em, valign=m]{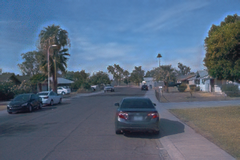}
			& \includegraphics[width=10em,height=5em, valign=m]{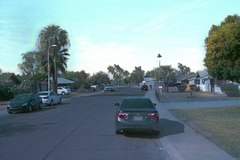}
			& \includegraphics[width=10em,height=5em, valign=m]{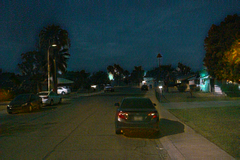}
			& \includegraphics[width=10em,height=5em, valign=m]{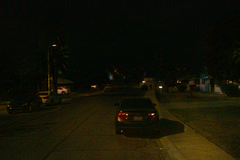}\\
            \midrule
			\adjustbox{valign=m}{\tiny\rotatebox{90}{\textbf{Few-shot}}}&\adjustbox{valign=m}{\rotatebox{90}{Ours}}
			& \includegraphics[width=10em, height=5em,valign=m]{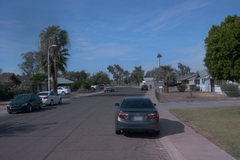}
			& \includegraphics[width=10em, height=5em,valign=m]{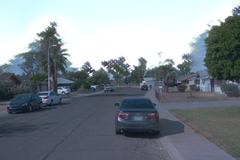}
			& \includegraphics[width=10em, height=5em,valign=m]{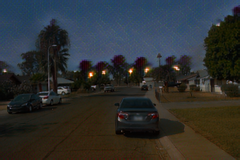}
			& \includegraphics[width=10em,height=5em, valign=m]{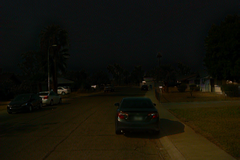}\\

		\end{tabular}\linkwithlabel{a}{b}{$w$}%
	}
	\caption{By enforcing multiple few-shot manifold deformations, we discover a realistic timelapse appearance comparable with CoMoGAN~\cite{pizzati2021comogan} and DNI-MUNIT~\cite{wang2019deep} (all non few-shot).} \label{fig:comogan}
\end{figure}

\section{Extensions}\label{sec:extensions}

\subsection{Few-shot continuous manifolds}

We investigate the use of \name for performing continuous image translation as in CoMoGAN~\cite{pizzati2021comogan}, thus learning the transformation from $\mathcal{S}{=}\textit{day}$ to $\mathcal{A}_m{=}\textit{night}$ on the Waymo~\cite{sun2020scalability} dataset by generating realistically intermediate frontal sun / twilight conditions where we have only few images. 
Here, we consider \textit{two} few-shot sets ($|\mathcal{T}|=10$), each one associated to one set of learned weights ($w^1, w^2$) between identity and night anchors. 
Results are in Fig.~\ref{fig:comogan}, where we also perform comparably to DNI-MUNIT~\cite{wang2019deep} and CoMoGAN which are trained with significantly more intermediate data (4721 vs 20). Please note that estimating $w^1$ and $w^2$ reorganizes the transformation realistically (\ie $\text{Day}\mapsto\text{Frontal sun}\mapsto\text{Twilight}\mapsto\text{Night}$) without prior knowledge on the order of few-shot set in the manifold. We evaluate mean rolling FID (mrFID) as in~\cite{pizzati2021comogan} and perform on par or better than baselines (for Model / StarGAN V2 / DNI - CycleGAN / DNI - MUNIT / CoMoGAN / \underline{Ours} we get 195 / 177 / 155 / 144 / 145 / \underline{145}). Only CoMoGAN (mrFID 137) outperforms thanks to its physical guidance.

\begin{figure}[t]
	\centering
	\includegraphics[width=\linewidth]{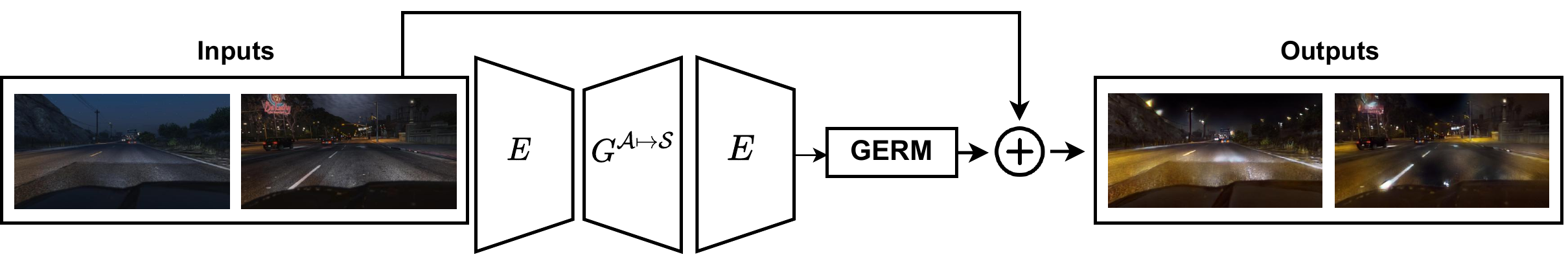}
	\caption{In the reconstruction cycle $\mathcal{A}_m\mapsto\mathcal{S}\mapsto\mathcal{A}_m$, we can inject the extracted residual with GERM on anchor images to perform an alternative  $\mathcal{A}_m\mapsto\mathcal{T}$ transformation.}
	\label{fig:wirediff}
\end{figure}

\subsection{Anchor-based translation}

The GERM extracts residual information from encoded source images. We investigate the application of residuals on the anchor images themselves, by first translating from $\mathcal{A}_m\mapsto\mathcal{S}$ using our backbone cycle consistency~\cite{huang2018multimodal}, and afterwards re-encoding the fake image in a $\mathcal{S}\mapsto\mathcal{A}_m$ reconstruction \textit{without retraining} (see Fig.~\ref{fig:wirediff}). 
This shows how \name simultaneously learns $\mathcal{S}\mapsto\mathcal{T}$ and acceptable $\mathcal{A}_m\mapsto\mathcal{T}$ transformations. The FID w.r.t. ACDC-Night improves from 142 to 130 when applying the residual on the synthetic anchors, thus confirming their shift towards $\mathcal{T}$.

\section{Conclusion}
\label{sec:discussion}
In this paper we presented \name, a framework for few-shot i2i which enables translating images to a single \textit{general} style approximating the entire few-shot set (\eg, for photo editing), or reproducing any specific \textit{exemplar} from the set for more variability (\eg, for domain adaptation). We demonstrated its effectiveness outperforming the state-of-the-art on many tasks, ablated its components and provided extensions to the framework.\smallskip

{
\noindent\textbf{Acknowledgements}: This work was partly funded by Vislab Ambarella, the French project SIGHT (ANR-20-CE23-0016), and received support from Service de coopération et d’action culturelle du Consulat général de France à Québec. It used HPC resources from GENCI–IDRIS (Grant 2021-AD011012808).
}

\clearpage
\bibliographystyle{splncs04}
\bibliography{egbib}
\end{document}